 \documentclass[preprint,3p,12pt]{elsarticle}

\usepackage[edges]{forest}
\usepackage{xcolor}
\usepackage[backref=false,breaklinks,colorlinks]{hyperref}  
\usepackage{adjustbox}

\usepackage{amssymb}
\usepackage{amsmath}
\journal{Information Fusion}

\begin{document}

\begin{frontmatter}

%% Title, authors and addresses

%% use the tnoteref command within \title for footnotes;
%% use the tnotetext command for theassociated footnote;
%% use the fnref command within \author or \affiliation for footnotes;
%% use the fntext command for theassociated footnote;
%% use the corref command within \author for corresponding author footnotes;
%% use the cortext command for theassociated footnote;
%% use the ead command for the email address,
%% and the form \ead[url] for the home page:
%% \title{Title\tnoteref{label1}}
%% \tnotetext[label1]{}
%% \author{Name\corref{cor1}\fnref{label2}}
%% \ead{email address}
%% \ead[url]{home page}
%% \fntext[label2]{}
%% \cortext[cor1]{}
%% \affiliation{organization={},
%%             addressline={},
%%             city={},
%%             postcode={},
%%             state={},
%%             country={}}
%% \fntext[label3]{}

\title{From Structure to Synergy: A Survey of Vision-Language Perception Paradigm Evolution in Multimodal Large Language Models} %% Article title

%% use optional labels to link authors explicitly to addresses:

\author[label1]{Haoxiang Sun}
\author[label1]{Tao Wang\corref{cor1}\textsuperscript{\dag}}
\author[label2]{Li Yuan}
\author[label3]{Jian Zhao}
\author[label1]{Jiancheng Lv}

\cortext[cor1]{\textsuperscript{\dag} Corresponding author. E-mail address: twangnh@gmail.com. \\Accepted by \textit{Information Fusion} on March 5, 2026, and currently in publication.}

\affiliation[label1]{organization={School of Computer Science, Sichuan University},%Department and Organization
            addressline={No.~24 South Section~1, Yihuan Road}, 
            city={Chengdu},
            postcode={610065}, 
            state={Sichuan},
            country={China}}

\affiliation[label2]{%
  organization={School of Electronic and Computer Engineering, Peking University Shenzhen Graduate School},
  addressline={2199 Lishui Road},
  city={Shenzhen},
  postcode={518055},
  state={Guangdong},
  country={China}
}
\affiliation[label3]{%
  organization={Institute of Artificial Intelligence (TeleAI), China Telecom and Northwestern Polytechnical University},
  addressline={127 Youyi West Road},
  city={Xi'an},
  postcode={710072},
  state={Shaanxi},
  country={China}
}

%% Author name
% + tao wang, li yuan, jian zhao, jianchenng lv

%% Author affiliation

%% Abstract
\begin{abstract}
%% Text of abstract
Multimodal Large Language Models (MLLMs) have recently made remarkable progress in unifying vision-language understanding and reasoning, especially following the introduction of models such as OpenAI’s O-series and DeepSeek’s R-series, which have driven a paradigm shift toward perception-centric intelligence. However, there remains a lack of systematic surveys that examine perception from a truly unified vision-language perspective—one that treats vision and language as an inseparable modality. Existing reviews are often fragmented, focusing separately on either vision or language, and thus rarely capture the cross-modal evolution of perception as an integrated capability.
To bridge this gap, we present the first systematic survey of unified vision-language perception in MLLMs. Specifically, we 
(1) formalize MLLM perception as an intrinsic, unified  vision-language capability analogous to human innate perception,
(2) introduce a five-stage taxonomy tracing the paradigm evolution of MLLM perception and survey representative methods and milestones at each phase, and
(3) identify open challenges and outline promising research directions toward truly general, unified multimodal intelligence.
We hope our study will provide both a foundational understanding and an actionable roadmap to foster further innovation on the path toward artificial general intelligence (AGI).
\end{abstract}

%%Graphical abstract
% \begin{graphicalabstract}
% %\includegraphics{grabs}
% \end{graphicalabstract}

% Research highlights
% \begin{highlights}
% \item Systematically reviews MLLM perception as a unified vision-language capability. 
% \item Proposes a five-stage taxonomy tracing the transition from structure to synergy.
% \item Outlines challenges and promising directions for future research.
% \end{highlights}

%% Keywords
\begin{keyword}
%% keywords here, in the form: keyword \sep keyword
Multimodal Large Language Models \sep Perception \sep Vision-Language Models \sep Unified Framework
%% PACS codes here, in the form: \PACS code \sep code

%% MSC codes here, in the form: \MSC code \sep code
%% or \MSC[2008] code \sep code (2000 is the default)

\end{keyword}

\end{frontmatter}

%% Add \usepackage{lineno} before \begin{document} and uncomment 
%% following line to enable line numbers
%% \linenumbers

%% main text
%%

%% Use \section commands to start a section
\section{Introduction}
\label{sec:introduction}
%% Labels are used to cross-reference an item using \ref command.

\begin{flushleft}
\leftskip=1cm\emph{``Semantic knowledge transforms the sensory cacophony into a symphony of meaning.''} \\
% , allowing us to recognize and make inferences about objects and events in the environment, and it provides the foundation for everyday behavioural acts.
\vspace{.3em}
\leftskip=4.55cm---\emph{Matthew A. Lambon Ralph}
\end{flushleft}
In recent years, Multimodal Large Language Models (MLLMs)\footnote{In this paper, “MLLM” is used as a unified term encompassing Large Language Models (LLMs), Large Vision-Language Models (LVLMs) and multimodal LLMs. Our focus is on their unified vision-language perception capabilities.} have evolved from a pure text-processing paradigm to unified vision-language understanding and reasoning, significantly advancing cross-modal tasks such as image captioning~\cite{wang2024tarsier,chai2024auroracap} visual grounding~\cite{yu2016modeling,lai2024lisa}, and multimodal question answering~\cite{liu2024mmbench,masry2022chartqa}.

The perception capability of MLLMs—the ability to extract, localize, and reason about structured semantic information from raw visual signals—serves as the cornerstone for  vision-language understanding and reasoning. It underpins applications ranging from classic computer vision benchmarks~\cite{yu2016modeling,lin2014microsoft} to open-ended, human-centric tasks~\cite{wu2024v,chen2024mega}.
This perception capability enables MLLMs to see, comprehend, and interact with multimodal information in a human-like manner~\cite{ralph2017neural}.

Despite remarkable advances and rapid evolution in MLLMs, there remains a lack of systematic surveys that approach these developments from the perspective of perception. Existing reviews are typically siloed, focusing on either vision-centric~\cite{sapkota2025object,shen2025reasoning} or language-centric~\cite{wang2025multimodal,chen2025towards} tasks, and thus fail to address perception as a unified vision-language capability. This fragmented perspective overlooks the paradigm-level transitions in vision-language perception within MLLMs, hindering a holistic understanding of their full potential and underscoring the need for an integrated, perception-focused review.

To address this gap, our survey provides the first in-depth and structured analysis of the evolution of\textit{ vision-language perception} in MLLMs. We clarify the scope of our review, outline its organization, and summarize our main contributions, aiming to serve as a roadmap for future research in unified vision-language perception.

\subsection{Scope and Definitions}

This section defines the scope of our survey, focusing on how MLLMs perceive images \emph{at the level of specific regions or instances in response to natural-language queries}.

Concretely, a method falls within our scope if it satisfies the following:
\begin{itemize}
    \item The input includes at least one natural 2D image or video clip.
    \item The core challenge lies in extracting or distinguishing visual evidence (objects, regions, attributes, relations), rather than in purely symbolic or mathematical reasoning. 
    \item Solving the task requires correctly perceiving specific regions or instances under natural-language guidance (e.g., referring expression comprehension, visual grounding, and region-level question answering).
\end{itemize}

Naturally, we \emph{explicitly exclude} three categories of work: 
(1) tasks primarily driven by logical or symbolic reasoning rather than perceptual processing (e.g., mathematical problem solving~\cite{cobbe2021training}); 
(2) methods targeting highly specialized applications in vertical domains (e.g., robotic vision--language control~\cite{walke2023bridgedata}); and 
(3) approaches whose gains arise solely from global architectural scaling (e.g., increasing input resolution~\cite{guo2024llava} or model parameter count~\cite{liu2024sphinx}) without introducing mechanisms that enhance localized perception.

\subsection{Paper Organization}
We categorize the evolution of perception capability into five stages: from structure-driven modular optimization (Stages I and II), to dynamic perception paradigms (Stage III), and ultimately to unified frameworks that synergize instruction, adaptivity, and reinforcement learning (Stage IVA and Stage IVB), offering prospects for the next generation of multimodal intelligence.
For an intuitive overview, the organizational framework of our survey is illustrated in Fig.~\ref{pic-perception-optimization}.

Specifically, the rest of this survey is organized as follows.
Sec. \ref{sec:preliminary} (Preliminary) introduces our five-stage taxonomy and contrasts our unified perception focus with prior, task-specific surveys. 
Sec. \ref{sec:stage1} (Stage I: Encoder-Centric Optimization) reviews early efforts to strengthen local perception via Region-Aware Modules and Integrated Perception Subnetworks. 
Sec. \ref{sec:stage2} (Stage II: Decoder-Centric Optimization) examines how auxiliary decoding shifts perception from region-level to pixel-level through Auxiliary Decoders, Multi-Decoder Architectures, and Specialized Decoding Strategies. 
Sec. \ref{sec:stage3} (Stage III: Dynamic Perception via Adaptive Processing) explores adaptive, context-driven perception across External Tool Scheduling, LLM-Centric Routing, and Code-as-Policy for Perception Tasks. 

Sec. \ref{sec:stage4a} (Stage IVA: Architecture-Free Strategies for Perception Enhancement) delves into two non-architectural approaches: Instruction-Based Strategies and RL-Based Strategies. 
Sec. \ref{sec:stage4b} (Stage IVB: Towards Unified Perception) provides a forward-looking perspective on the convergence of instruction, adaptivity, and RL via Why Unification Matters, Signs of Convergence and Towards a New Generation of Perception-Centric Agents. 
Finally, Sec. \ref{sec:conclusion} offers a concise summary of our key findings and outlines directions for future research.

\subsection{Main Contributions}
\begin{itemize}
\item \textbf{First systematic survey:} To the best of our knowledge, this is the first survey to systematically analyze vision-language perception in MLLMs as an intrinsic, unified capability, providing an actionable roadmap for future research.
\item \textbf{Paradigm taxonomy and comprehensive review:} We propose a five-stage taxonomy of paradigm evolution in MLLM perception and comprehensively review representative works at each stage, highlighting their key innovations.
\item \textbf{Emerging challenges and future directions:} We provide an in-depth discussion of current challenges and outline promising directions for future development in perception-centric multimodal intelligence.
\end{itemize}

\section{Preliminary}
\label{sec:preliminary}
\subsection{Taxonomy}
\label{subsec:taxonomy}
In this section, we introduce our five-stage taxonomy that characterizes the evolutionary trajectory of perception capabilities in MLLMs. This framework is designed to capture the paradigm shift from structure—with targeted, architecture-driven enhancements—to synergy, where perception emerges from integrated, adaptive, and unified strategies.

Our taxonomy begins with encoder-centric, structure-driven approaches developed for specific scenarios such as simple image understanding tasks in Visual Question Answering (VQA), where perception improvements are achieved either by enhancing the internal structure of the vision encoder or by modifying the model architecture to treat the encoder as an integrated perceptual subnetwork. It then moves to decoder-centric strategies, which further advance fine-grained perception by employing specialized decoding designs that increase the granularity of the model’s perceptual capabilities—from region-level to pixel-level understanding. The third stage, dynamic perception, focuses on context-aware and adaptive processing, enabling models to flexibly interact with images multiple times to acquire richer and more comprehensive information.

The most recent advances represent two distinct but related directions: architecture-free strategies, which focus on optimizing perception for specific tasks without explicit architectural changes; and unified perception frameworks, which emphasize the model’s ability to operate in complex, open-ended scenarios by integrating instruction, adaptivity, and reinforcement learning. Despite their different emphases, both directions—like the earlier stages—continue to advance local perception capabilities as a central theme.

\begin{figure}[h]
    \centering
    \includegraphics[width=1\linewidth]{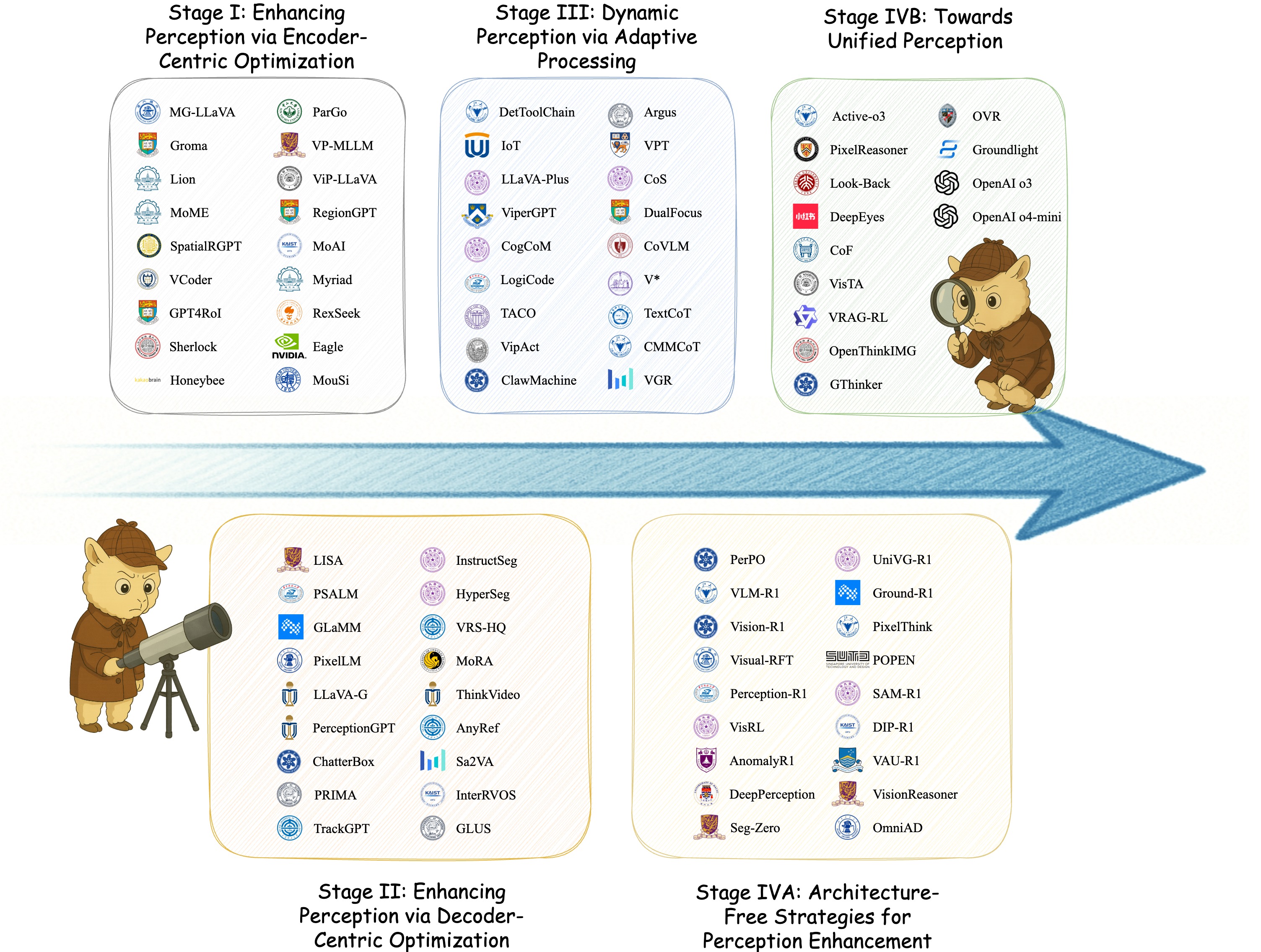}
    \caption{ An overview of the evolving paradigms for enhancing  perception in multimodal large language models. The arrow indicates the overall trajectory of evolution, moving from structural modifications to synergy-driven integration.}
    \label{fig:overview}
\end{figure}

Importantly, our taxonomy should not be viewed as a strictly linear progression. Many of these stages are intrinsically intertwined, with considerable overlap and mutual influence between them. We categorize each representative work according to its primary contribution to the evolution of perception in MLLMs, rather than its chronological order, to better reflect the conceptual landscape and research focus of the field. 
To provide a clear and intuitive overview, we illustrate these five stages and their relationships in Fig.~\ref{fig:overview}.

\subsection{Differences from Related Work}
\label{subsec:diff-related-work}
% Recently, most surveys in this area have concentrated on two main directions: one branch focuses on specific vision tasks~\cite{sapkota2025object,shen2025reasoning}, while the other emphasizes symbolic and mathematical reasoning~\cite{wang2025multimodal,chen2025towards,yan2024survey,zhao2025towards}. These works have certainly contributed to advancing the field. However, the former tends to emphasize isolated vision tasks while overlooking the holistic perceptual capabilities inherent to MLLMs, whereas the latter prioritizes abstract language-based reasoning, often neglecting the visual reasoning and fine-grained perceptual understanding that are central to our investigation. As a result, the community still lacks a comprehensive understanding of how MLLMs jointly develop and evolve their perception abilities across modalities.

% In contrast, our survey adopts a unified, perception-centered perspective on understanding and reasoning in MLLMs. We systematically analyze the evolution of perception in MLLMs—tracing the progression from structure-driven modular optimization, through dynamic and context-aware processing, to unified frameworks that integrate instruction tuning, adaptivity, and reinforcement learning. This perspective reframes each developmental stage as a step toward more human-like, perception-centric  vision-language reasoning.

Recent surveys have already reviewed multimodal large language models from multiple perspectives, including architectures, evaluation, perception, and reasoning. Our work is most closely related to these efforts but \textit{differs in both scope and organizing perspective}. In short, they can be grouped into the following categories:

\begin{itemize}
    \item Comprehensive surveys on the evolution of MLLMs: 
    Liang et al.~\cite{liang2024comprehensive} provide a comprehensive guide to MLLMs and examine their architectures, applications, and impact, covering topics such as training methods, architectural components, and practical applications in various fields. 
    Caffagni et al.~\cite{caffagni2024revolution} review earlier stages of MLLM development and analyze their architectural choices, multimodal alignment strategies, and training techniques. 
    Similarly, Zhang et al.~\cite{zhang2024mm} review the evolution of MLLMs over time, focusing on the performance of selected models on mainstream benchmarks and summarizing key training recipes. 
    These surveys chart the overall evolution of MLLMs from a broad, time-oriented perspective. In contrast, our work restricts the scope to the narrower problem of vision–language perception, with a particular focus on region- or instance-level understanding under natural-language guidance.
    \item Task-specific surveys: Sapkota et al.~\cite{sapkota2025object} present an in-depth review of object detection with MLLMs, analyzing how these models leverage natural language and visual features to perform object localization and category recognition, and comparing their adaptability and efficiency to conventional deep detectors. Shen et al.~\cite{shen2025reasoning} focus on reasoning segmentation for images and videos, summarizing existing methods, benchmarks, and applications across diverse domains. These surveys are all centered on specific vision tasks and, although illustrative, do not capture how perception modules evolve toward synergistic improvements within a unified vision-language paradigm.

    \item Chain-of-thought and mathematical reasoning surveys: Wang et al.~\cite{wang2025towards} present a survey of multimodal chain-of-thought(MCoT) reasoning, clarifying definitions, summarizing related MCoT methods, and organizing them by application scenarios. Chen et al.~\cite{chen2025towards} focus on long chain-of-thought for reasoning LLMs, proposing a taxonomy that distinguishes long versus short CoT, analyzing phenomena such as overthinking and test-time scaling, and discussing future directions including multimodal extensions. These works take reasoning as the primary organizing axis and largely treat visual signals as contextual input, whereas our survey centers on how vision-language perception itself evolves. Zhou et al.~\cite{zhou2025perception} deconstruct vision-language interactive reasoning into a foundational Perception layer for accurate visual information extraction and fine-grained alignment, and a higher-order Cognition layer for proactive, multi-step, goal-oriented reasoning built upon this perceptual basis. Guided by this two-layer framework, their survey analyzes bottlenecks and methods at both layers, with a primary focus on the hierarchical relationship between visual inputs and logical reasoning within interactive observe–think–verify loops. In contrast, our survey adopts an evolutionary, perception-centered perspective: we concentrate on paradigm shifts in perception architectures themselves, tracing the trajectory from modular, structure-driven designs to a unified, synergy-driven vision-language perception paradigm.

\end{itemize}

Taken together, existing surveys either provide broad overviews of MLLMs, offer vertical reviews of individual perception tasks, or organize the landscape around chain-of-thought and interactive reasoning. By contrast, our survey treats vision–language perception itself as the central organizing axis: we precisely restrict the scope to region- and instance-level perception under natural-language guidance and trace its paradigm shifts across five stages, from encoder/decoder-centric designs to dynamic and architecture-free strategies, culminating in unified, synergy-driven vision–language perception.

\definecolor{color-layer2}{HTML}{000000}
\definecolor{color-layer1}{HTML}{319D9C}
\definecolor{color-layer0}{HTML}{C6AF3C}
\definecolor{color-layer3}{HTML}{A5BED5}

\tikzstyle{leaf-style}=[
    rectangle,
    draw=none,
    rounded corners,
    text opacity=1,
    minimum height=1.5em,
    minimum width=20em,
    inner sep=2pt,
    align=left,
    fill opacity=.5,
    line width=0.9pt,  
]
\tikzstyle{leaf}=[leaf-style, minimum height=1.5em,
    fill=color-layer3!30, text=black, align=left,font=\scriptsize,
    inner xsep=2pt,
    inner ysep=2pt,
    line width=0.8pt,
    align=left,        % ← 覆写全局的 align=center
    text ragged,       % ← 让换行后也左对齐而不是两端对齐
    anchor=west,       % ← 文字锚点对齐到节点左边
]
\tikzset{
  leaf2/.style={
    % 先继承 leaf
    leaf,
    % 再覆盖 inner xsep
    inner xsep=0pt,
    align=center,
    % （你还可以在这里改 inner ysep、align 等）
  }
}

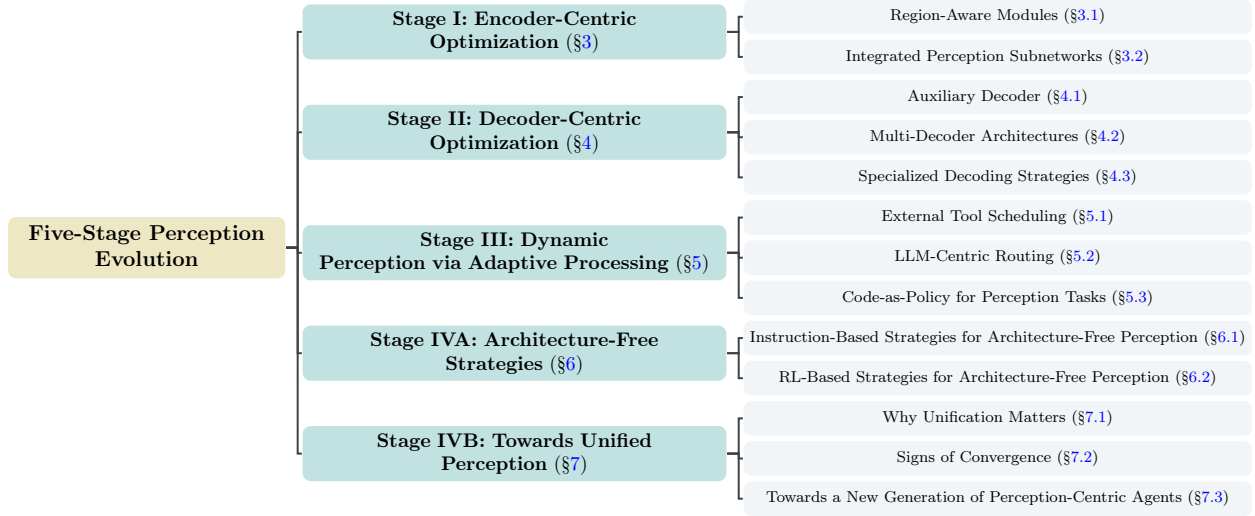
\begin{figure}
    \centering
    \begin{adjustbox}{max width=\linewidth}
        \begin{forest}
            forked edges,
            for tree={
                grow=east,
                reversed=true,
                anchor=base west,
                parent anchor=east,
                child anchor=west,
                base=center,
                font=\small,
                rectangle,
                draw=none,
                rounded corners,
                align=center,
                text centered,
                minimum width=4em,
                edge+={darkgray, line width=1pt},
                s sep=2pt,
                l sep=8pt,
                inner xsep=2pt,
                inner ysep=2pt,
                line width=0.8pt,
                ver/.style={rotate=0, child anchor=west, parent anchor=east, anchor=center,minimum width=12em},
            },
            where level=1{text width=18em,font=\footnotesize,}{},
            where level=2{text width=22em,font=\scriptsize,}{},
            where level=3{text width=16em,font=\scriptsize,}{},
            [
                \textbf{Five-Stage Perception}\\\textbf{Evolution}, ver, fill=color-layer0!30
                [
                    \textbf{Stage I: Encoder-Centric}\\\textbf{Optimization} (\S\ref{sec:stage1}), fill=color-layer1!30
                    [
                        Region-Aware Modules (\S\ref{subsec:region-aware}), leaf2
                    ]
                    [
                        Integrated Perception Subnetworks (\S\ref{subsec:integrated-subnetworks}), leaf2
                    ]
                ]
                [
                    \textbf{Stage II: Decoder-Centric}\\\textbf{Optimization} (\S\ref{sec:stage2}), fill=color-layer1!30
                    [
                        Auxiliary Decoder (\S\ref{subsec:auxiliary-decoder}),leaf2
                    ]
                    [
                        Multi-Decoder Architectures (\S\ref{subsec:multi-decoder-arch}), leaf2
                    ]
                    [
                        Specialized Decoding Strategies (\S\ref{subsec:specialized-decoding}), leaf2
                    ]            
                ]
                [
                    \textbf{Stage III: Dynamic}\\\textbf{Perception via Adaptive Processing} (\S\ref{sec:stage3}), fill=color-layer1!30
                    [
                        External Tool Scheduling (\S\ref{subsec:external-tool}),leaf2
                    ]
                    [
                        LLM-Centric Routing (\S\ref{subsec:llm-centric-routing}), leaf2
                    ]
                    [
                        Code-as-Policy for Perception Tasks (\S\ref{subsec:code-as-policy}), leaf2
                    ]
                ]
                [
                    \textbf{Stage IVA: Architecture-Free}\\\textbf{Strategies} (\S\ref{sec:stage4a}), fill=color-layer1!30
                    [
                        Instruction-Based Strategies for Architecture-Free Perception (\S\ref{subsec:instruction-based}), leaf2
                    ]
                    [
                        RL-Based Strategies for Architecture-Free Perception (\S\ref{subsec:rl-based}), leaf2
                    ]
                ]
                [
                    \textbf{Stage IVB: Towards Unified}\\\textbf{Perception} (\S\ref{sec:stage4b}), fill=color-layer1!30
                    [
                        Why Unification Matters (\S\ref{subsec:why-unification-matters}), leaf2
                    ]
                    [
                        Signs of Convergence (\S\ref{subsec:signs-of-convergence}), leaf2
                    ]
                    [
                        Towards a New Generation of Perception-Centric Agents (\S\ref{subsec:new-gen-perception-agents}), leaf2
                    ]
                ]
            ]
            \end{forest}
        \end{adjustbox}
    \caption{Organization of our five-stage evolutionary framework.}
    \label{pic-perception-optimization}
\end{figure}

\section{Stage I: Enhancing Perception via Encoder-Centric Optimization}
\label{sec:stage1}
\begin{figure}[h]
    \centering
    \includegraphics[width=1\linewidth]{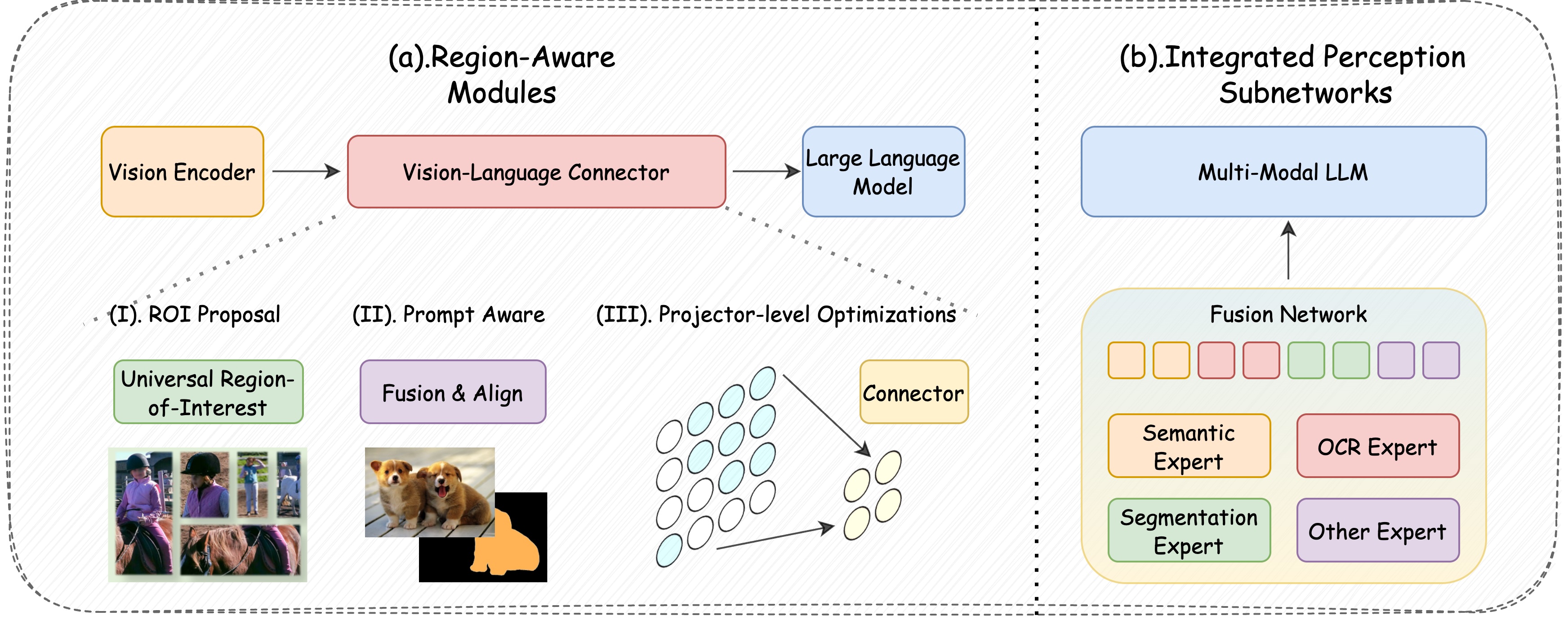}
    \caption{Encoder-Centric Optimization Strategies for Multimodal Perception in MLLMs.  (a) Region-aware modules enhance perceptual capability by generating universal ROI proposals within the encoder, performing post-encoding feature fusion and alignment, or applying task-specific optimizations at the projector/connector level to better adapt the model to particular visual contexts. (b) Integrated perception subnetworks achieve multi-level information fusion by incorporating multiple visual expert subnetworks directly within the encoder stage.}
    \label{fig:stage1}
\end{figure}
Stage I focuses on enhancing perception by modifying the encoder itself. Early works pursue encoder-internal optimizations, introducing components such as region-aware modules and projector- or connector- level adjustments to strengthen fine-grained, local perception (\S\ref{subsec:region-aware}). In parallel, another line of research treats the encoder as an integrated perception subnetwork, aiming to improve perceptual capacity by either expanding the variety of encoders or composing multiple specialized encoder branches (\S\ref{subsec:integrated-subnetworks}). As provided in Fig.~\ref{fig:stage1}, these strategies together elevate MLLMs’ perceptual granularity from an image-level view to region-level understanding, laying the groundwork for more advanced paradigms that extract, localize, and reason about visual information with finer precision.
\subsection{Region-Aware Modules}
\label{subsec:region-aware}

A central challenge in multimodal perception is enabling models to selectively attend to and process meaningful regions within an image, rather than relying solely on global representations. Region-aware modules address this by equipping the encoder with mechanisms to enhance local, fine-grained perception. We group region-aware modules into three parts: Universal Region-of-Interest Proposal (\S\ref{subsubsec:universal roi}), Query- or Prompt-Aware Encoders (\S\ref{subsubsec:Query- or Prompt}), Projector-level or Connector-level Optimizations (\S\ref{subsubsec:projector-level}).
\subsubsection{Universal Region-of-Interest Proposal}
\label{subsubsec:universal roi}
Some works propose universal region-of-interest (ROI) generation methods~\cite{he2017mask} that are decoupled from the large language model, thereby enabling more flexible and generalizable region selection, independent of task-specific semantics. 
MG-LLaVA~\cite{zhao2024mg} designs a multi-granularity encoder with separate pathways for different resolutions and object-level ROIs, with the object-level ROI generation incorporating both a tagging model~\cite{zhang2024recognize} and an open-vocabulary detector~\cite{minderer2023scaling}.
ChatRex~\cite{jiang2024chatrex} integrates a universal proposal network~\cite{he2017mask} within the MLLM, projecting regression-predicted boxes into the LLM and reframing box prediction as a retrieval task. 
Similarly, Groma~\cite{ma2024groma} integrates a Deformable DETR~\cite{zhu2020deformable} as its core region proposal module, and combines this with specialized region tokens to achieve localized visual tokenization.
Apart from that, some works focus on specific scenarios: 
ASM~\cite{wang2024all} performs ROI selection for panoptic visual recognition, followed by a location-aware tokenizer to encode selected regions.
Artemis~\cite{qiu2024artemis} proposes an ROI tracking and selection mechanism for video-based referring, obtaining a list of boxes that represent spatiotemporally aware ROIs.

\subsubsection{Query- or Prompt-Aware Encoders}
\label{subsubsec:Query- or Prompt}
Several approaches enhance region awareness by making region selection adaptive to various prompts—whether language- or vision-based—thus enabling more context-sensitive and interactive perception.

Some works focus specifically on visual prompts, designing dedicated modules to encode or fuse visual region cues as part of the input.
To handle visual prompts at the mask level, 
Osprey~\cite{yuan2024osprey} and Finecaption~\cite{hua2025finecaption} use a Mask-Aware Visual Extractor that ingests both the input image and referring masks to obtain mask-region perception.  
VP-MLLM~\cite{lin2024draw} introduces a  Visual Prompt Encoder to accommodate and recognize various types of visual prompts as input. 

Moving toward free-form visual prompts, Ferret~\cite{you2023ferret} introduces a Spatial-Aware Visual Sampler that enables the acquisition of visual features from regions of arbitrary shape, accommodating varying spatial sparsity and thus supporting the processing of free-form region inputs. 
VPP-LLaVA~\cite{tang2025visual} introduces two complementary mechanisms: a global Visual Position Prompt, which overlays a learnable, axis-like tensor onto the input image to provide structured spatial cues, and a local Visual Position Prompt, which incorporates position-aware queries to enable fine-grained localization.

Beyond these specialized vision-aware modules, 
RegionGPT~\cite{guo2024regiongpt} employs dual encoder branches—patch-merge for image-level features and mask-pooling for region-level features—to improve region specificity, while
GPT4ROI~\cite{zhang2025gpt4roi} refines spatial instruction tuning by interleaving spatial instructions and explicit region references within a single token sequence, combining RoIAlign-extracted region features with language embeddings for precise region localization.
In contrast with the above approaches that design dedicated modules for handling visual prompts, ViP-LLaVA~\cite{cai2024vip} forgoes any extra modules and directly fuses visual prompts with the original image within the vision encoder using an alpha-blending mechanism, thus supporting free-form visual prompt integration.

Other works concentrate on language prompts. 
Lion~\cite{chen2024lion} integrates a Recognize Anything Model (RAM)~\cite{zhang2024recognize} for extracting image tags as soft prompts and employs a Mixture-of-Adapters with a router mechanism. The router dynamically fuses image-level or region-level features from different visual branches, enabling adaptive perception across diverse scenarios.
FlexCap~\cite{dwibedi2024flexcap} targets the captioning task by accepting object-box coordinates and directly linearly projecting them into the LLM during training. 

In addition, some works~\cite{liu2023visual,peng2023kosmos} enhance fine-grained perception without modifying the encoder architecture itself, but rather through instruction tuning or prompt engineering. We discuss these strategies in Stage IVA (\S\ref{sec:stage4a}).

\subsubsection{Projector-level or Connector-level Optimizations}
\label{subsubsec:projector-level}
Other methods achieve region awareness via projector-level or connector-level optimizations. Honeybee~\cite{cha2024honeybee} proposes a novel projector design that is both flexible and locality-enhanced. ParGo~\cite{wang2025pargo} introduces a Partial-Global projector that aligns two separately pre-trained models by integrating partial and global views, mitigating overemphasis on prominent regions.

Connector‑level approaches have likewise been developed to bridge vision encoders and LLMs:  
Groundhog~\cite{zhang2024groundhog} enhances holistic segmentation by employing a Masked Feature Extractor as a connector, together with a Mask Proposal Model and a Mask Retrieval Head. It treats segmentation as a retrieval task over class-agnostic entity mask proposals, thereby improving perceptual performance.
For video streams, 
CountLLM~\cite{yao2025countllm} inserts a periodicity transformer between the video encoder and LLM to achieve periodicity‑aware alignment for repetitive action counting; Elysium~\cite{wang2024elysium} employs a T‑Selector module to enable the MLLM to process a larger number of frames without significant performance degradation; and TimeChat~\cite{ren2024timechat} designs a Sliding Video Q‑Former with a Time‑Aware Frame Encoder that first extracts spatial tokens per frame, binds them with timestamp descriptions, and then uses a moving sliding window to establish temporal relations across varied‑length video tokens.

\subsection{Integrated Perception Subnetworks}
\label{subsec:integrated-subnetworks}
Beyond internal encoder modifications, another line of research treats the encoder as an integrated perception subnetwork. These approaches aim to enhance perceptual capacity by either increasing the number and diversity of encoder branches (\S\ref{subsubsec:specialized encoders}), or by composing multiple specialized encoders within the model architecture (\S\ref{subsubsec:Multiple visual modal}).

\subsubsection{Specialized Task Encoders}
\label{subsubsec:specialized encoders}
Some works focus on enhancing perception by introducing multiple specialized encoders, each tailored for specific task types. Eagle~\cite{shi2024eagle}, MouSi~\cite{fan2024mousi} and MoME~\cite{shen2024mome} employ a mixture of specialized vision encoders, introducing distinct vision experts for different scenarios such as OCR, detection, and segmentation. Visual features produced by these experts are fused and then projected into the MLLM, allowing the model to adaptively leverage multiple visual perspectives. 
Other works target more specific scenarios: Sherlock~\cite{ma2025sherlock} addresses video anomaly perception by introducing a Global-Local Spatial-Enhanced Mixture of Expert  module with a Spatial Imbalance Regulator. The former module includes four spatial experts for extracting spatial information and an expert gate to balance global and local spatial cues, effectively tackling the challenge of global-local spatial modeling. 
Similarly, Points~\cite{liu2024points} employs a dual-encoder branch that fuses representations from an OCR module and a general module.
RexSeek~\cite{jiang2025referring} builds upon ChatRex~\cite{jiang2024chatrex} by integrating a person detection model.

Building on these multi-expert encoders, other works introduce auxiliary branches to incorporate external vision modules alongside the primary encoder. 
MoAI~\cite{lee2024moai} introduces a bypass branch for the vision encoder, directly invoking conventional computer vision models to process the input image. The results are verbalized and compressed using a dedicated compressor before being embedded into the MLLM.  
Myriad~\cite{li2023myriad} leverages pre-trained industrial anomaly detection (IAD) models to generate anomaly maps, which are fed into a specially designed vision encoder.
ROD-MLLM~\cite{yin2025rod} introduces an additional open-vocabulary detection (OVD) module to recall candidate objects for ROI alignment.

Moreover, IVE~\cite{he2024incorporating} combines both strategies: it calls specific visual tools to extract structured data as hard prompts and utilizes multi-task encoders to obtain richer visual information, aggregating available visual information through a mixture-of-experts mechanism in both training and inference stages.

\subsubsection{Multiple Visual-modal Encoders}
\label{subsubsec:Multiple visual modal}
A related direction enriches perception by integrating specialized visual modalities (e.g., depth, segmentation) within the encoder architecture. SpatialRGPT~\cite{cheng2024spatialrgpt} introduces a flexible plugin module for 3D spatial awareness, combining masks/boxes feature extractors with RGB and depth connectors to capture spatially diverse information. Similarly, VCoder~\cite{jain2024vcoder} fuses multiple perception modalities during encoding, yielding richer and more comprehensive region representations.

\section{Stage II: Enhancing Perception via Decoder-Centric Optimization}
\label{sec:stage2}
\begin{figure}[h]
    \centering
    \includegraphics[width=1\linewidth]{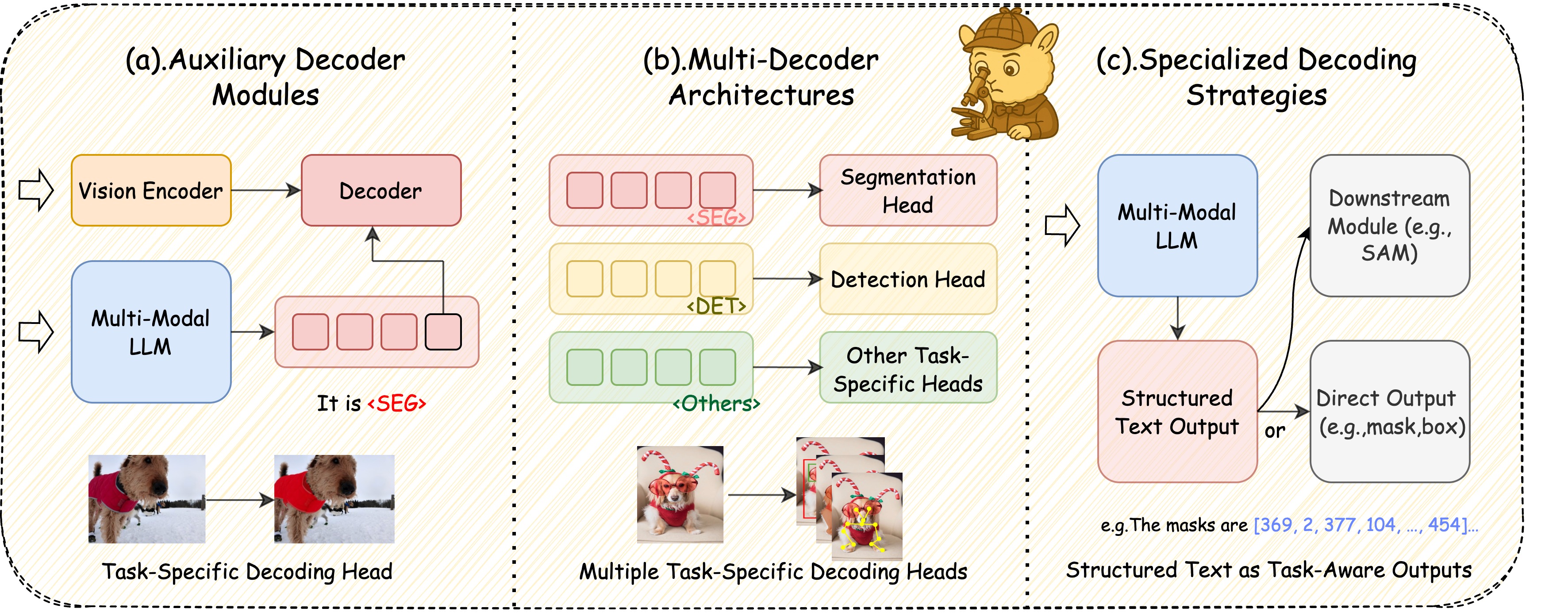}
    \caption{Decoder‐Centric Optimization Strategies for Multimodal Perception in MLLMs.  
(a) Auxiliary decoder modules enhance perception by integrating task-specific decoding heads alongside vision encoders.  
(b) Multi-decoder architectures employ multiple parallel task-specific heads (e.g., segmentation, detection, and others) to support diverse perception tasks.  
(c) Specialized decoding strategies leverage structured text outputs, which can either guide downstream modules (e.g., SAM) or serve as direct task outputs (e.g., masks or boxes).}
    \label{fig:pic_decoder}
\end{figure}

While encoder modifications have advanced MLLMs to capture finer-grained region-level features, their analysis typically remains at the region level, limiting precise localization  capabilities. Stage II marks a pivotal shift: by introducing auxiliary decoders, MLLMs make a substantial leap from region-level to true pixel-level perception. Although these decoders differ in their architectures and target tasks, they can be viewed as proxy objectives that leverage the MLLM’s vision-language backbone to adapt to diverse, task-specific requirements—thereby enhancing both specialized and general perception capabilities. As shown in Fig.~\ref{fig:pic_decoder}, we organize decoder-centric approaches into three categories: Auxiliary Decoder (\S\ref{subsec:auxiliary-decoder}), Multi-Decoder Architectures (\S\ref{subsec:multi-decoder-arch}), and Specialized Decoding Strategies (\S\ref{subsec:specialized-decoding}).

\subsection{Auxiliary Decoder}
\label{subsec:auxiliary-decoder}

This subsection surveys auxiliary decoder approaches that advance MLLM perception from region-level analysis to fine-grained pixel-level understanding. 
In the following, we detail these advances in three parts: Early Auxiliary Decoder: The LISA Paradigm (\S\ref{subsubsec:lisa-auxiliary}), Temporal and Multi-Image Auxiliary Decoder (\S\ref{subsubsec:temporal-multiimage-auxiliary}), and Auxiliary Decoder for Multi-Granularity Perception (\S\ref{subsubsec:multi-granularity-decoder}).

\subsubsection{Early Auxiliary Decoder: The LISA Paradigm}
\label{subsubsec:lisa-auxiliary}

A foundational line of early work introduces segmentation as an auxiliary decoding task in MLLMs, with LISA~\cite{lai2024lisa} serving as a pioneering example. LISA enables the MLLM to generate a special $\langle seg \rangle$ token, whose last-layer embedding is decoded into a pixel-level segmentation mask using the Segment Anything Model (SAM)~\cite{kirillov2023segment}. This marks the first end-to-end segmentation capability within the MLLM framework.

Building on this foundation, a growing body of work has extended the auxiliary decoder paradigm in multiple directions. One prominent line of research focuses on enhancing multimodal input capabilities. GLaMM~\cite{rasheed2024glamm} builds on LISA by incorporating multiple encoders—global, regional, and grounding image encoders—thereby supporting both textual and visual prompts as well as multi-region pixel-level grounding. PerceptionGPT~\cite{pi2024perceptiongpt} follows a similar route, using lightweight visual encoders and decoders to project visual features into the LLM’s embedding space as discrete tokens, allowing visual information to be handled alongside text. 
OMG-LLaVA~\cite{zhang2024omg} employs a universal perception module~\cite{li2024omg} as its visual encoder, integrating image information, perception priors, and visual prompts into unified visual tokens for the LLM. 
AnyRef~\cite{he2024multi} proposes a Unified Referring Representation method to encode references from multiple modalities into LLM-aligned embeddings. A Refocusing Mechanism is further employed to enrich these embeddings with grounded textual cues, offering enhanced representational power.

Another thread of research seeks to broaden the task scope of segmentation.  PSALM~\cite{zhang2024psalm} expands on LISA’s framework by introducing a conditional prompt: the output embedding of this prompt serves as classifier weights for predicting segmentation mask categories. This approach
enables the model to generalize from referring segmentation task to various  segmentation
tasks. 
ChatterBox~\cite{tian2024chatterbox} targets multi-round referring and grounding tasks by using the $\langle gnd \rangle$ token as an LLM query to guide DINO~\cite{zhang2022dino} in decoding object locations.

In parallel, several efforts aim to enhance localization without compromising the original dialogue and reasoning abilities of the MLLM. 
LISA++\cite{yang2023lisa++} enhances reasoning instance segmentation ability and
more natural text responses with the ability of Segmentation in Dialogue. 
LLaVASeg~\cite{yang2024empowering} employs a chain-of-thought prompting strategy with lightweight adapters to instruct MLLMs to segment user-specified target regions, maintaining their dialogue capabilities while equipping them with strong reasoning-driven segmentation ability. 
Similarly, MIRSA~\cite{cai2025pixel} combines segmentation with multi-turn interaction, 
along with LLM-based reasoning quality  evaluation metrics. 
SegLLM~\cite{wang2024segllm} addresses multi-round interactive image reasoning segmentation by introducing a mask-aware decoding scheme and four special tokens, which enables the model to generate new masks based on both current LLM outputs and the memory of previous masks.

Beyond expanding capability and versatility, some works pursue enhanced segmentation accuracy and efficiency. PixelLM~\cite{ren2024pixellm} proposes a lightweight decoder and a comprehensive segmentation codebook. The decoder utilizes the hidden embeddings of codebook tokens, encoding detailed and target-specific information to produce segmentation masks quickly and accurately within the MLLM.
ROSE~\cite{han2024rose} targets open-set dense segmentation by treating each image patch as an independent region-of-interest candidate. This design enables the model to perform patch-wise perception, mask and category decoding, thereby achieving both dense and sparse mask predictions. 
GSVA~\cite{xia2024gsva} augments the decoder with multiple $\langle seg \rangle$ tokens and a dedicated $\langle rej \rangle$ token, allowing the model to explicitly reject null targets. 
Similarly, LaSagnA~\cite{wei2024lasagna} employs both $\langle seg \rangle$ and $\langle rej \rangle$ tokens to effectively handle complex queries.

\subsubsection{Temporal and Multi-Image Auxiliary Decoder}
\label{subsubsec:temporal-multiimage-auxiliary}
Some works extend the auxiliary decoder beyond single-frame inputs to sequences or multi-image contexts, enabling pixel-level segmentation and tracking across time and views. 

A strand of research pursues universal segmentation across images and videos by explicitly extracting temporal structure from video. 
InstructSeg~\cite{wei2024instructseg} employs an object-aware video perceiver to extract temporal and object cues from reference frames and uses vision-guided multi-granularity text fusion to integrate global and detailed prompt information, enabling both referring and reasoning segmentation across images and videos.  
VRS-HQ~\cite{gong2025devil} introduces Temporal Dynamic Aggregation and token-driven keyframe selection, and augments the conventional $\langle seg \rangle$ token with an additional temporal $\langle tak \rangle$ token.
ThinkVideo~\cite{kao2025thinkvideo} introduces a multi-agent, zero-shot framework that leverages the chain-of-thought (CoT) reasoning abilities of MLLMs for enhanced video segmentation. The method extracts object selectivities for keyframes and links a reasoning segmentation model with the SAM2~\cite{ravi2024sam} video processor to produce mask sequences.
ViLLa~\cite{zheng2024villa} employs a key segment extractor, context synthesizer, and hierarchical temporal synchronizer to align video-level and frame-level segmentation tokens, producing coherent multi-frame segmentations.
VISA~\cite{yan2024visa} introduces a text-guided frame sampler to select the most distinctive frame as the segmentation target, along with corresponding reference frames.

A complementary direction seeks to cast heterogeneous tasks into a single interaction paradigm. 
Sa2VA~\cite{yuan2025sa2va} unifies various tasks—spanning static and dynamic visual understanding—under a single instruction-tuning process, encoding all inputs (text, visual prompts, images, and videos) as token embeddings for comprehensive multimodal grounding.
HyperSeg~\cite{wei2025hyperseg} augments an MLLM with a universal segmentation framework that couples a hybrid entity-recognition module with a fine-grained visual perceiver. With an additional temporal adapter, the framework extends to challenging video tasks.

Distinct from the two branches above, another line of work tailors auxiliary decoders to video-specific scenarios. 
TrackGPT~\cite{zhu2023tracking} integrates segmentation into tracking by introducing a self-correcting rethinking  mechanism that revises predictions deviating from the intended instruction, and a cross-frame referring propagation module that leverages cues from adjacent frames to ensure robust and accurate tracking.
MoRA~\cite{deng2025motion} targets motion-grounded video reasoning by introducing a $\langle loc \rangle$ token for temporal information embedding and a temporal localization head that decodes binary temporal masks to refine the raw outputs from SAM.  
ReVIOSa~\cite{jin2025interrvos} targets interaction-aware referring video object segmentation and introduces two special tokens, $\langle seg\_act \rangle$ and $\langle seg\_tar \rangle$, to separately segment the referred subject and the interacting object.
GLUS~\cite{lin2025glus} integrates global and local reasoning into a single MLLM for referring video object segmentation with pre-trained memory modules, and introduces plug-and-play self-refinement via key-frame selection and object-contrastive loss. 

Beyond videos, PRIMA~\cite{wahed2024prima} targets multi-image pixel-grounded reasoning by combining a DINOv2-based vision encoder for dense semantics with a Q-Former’s selective cross-attention, dynamically generating masks for objects and parts referenced in natural language queries. 
CALICO~\cite{nguyen2025calico} addresses part-focused semantic co-segmentation across multiple images by integrating a Correspondence Extraction Module, which captures semantically rich part correspondences, and Correspondence Adaptation Modules, which inject this information into the MLLM’s representation to enable efficient and accurate co-segmentation.

\subsubsection{Auxiliary Decoder for Multi-Granularity Perception}
\label{subsubsec:multi-granularity-decoder}

Beyond conventional single-scale segmentation, several approaches have introduced auxiliary decoders designed for multi-granularity perception, further enhancing the ability of MLLMs to process both coarse global context and fine local details.
UniRES++~\cite{liu2025towards} is among the first MLLMs explicitly designed for multi-granularity referring expression segmentation. It features a Multi-Granularity Vision Flow for capturing multi-level visual features, a Grounding Encoder for foundational representations, an LLM backbone, dynamic Multi-Granularity Feature Exploitation for adaptive feature selection, and a Pixel Decoder for segmentation mask generation.
Similarly, $M^2\text{SA}$~\cite{jang2025mmr} addresses multi-target, object-level, and part-level reasoning segmentation by introducing early local feature fusion and employing multiple $\langle seg \rangle$ tokens. 
Likewise, MGLMM~\cite{yuan2025instruction} introduces multiple $\langle seg \rangle$ tokens along with an additional projector to better align image features with the linguistic modality, enabling seamless switching between multi-granularity segmentation and captioning tasks.

\subsection{Multi-Decoder Architectures}
\label{subsec:multi-decoder-arch}
In this line of work, the MLLM serves as a central backbone, equipped with multiple task-specific decoders to accomplish diverse vision tasks within a unified framework.

NExT-Chat~\cite{zhang2023next} and u-LLaVA~\cite{xu2024u} are two early efforts exploring multi-task perception within MLLMs. Specifically, NExT-Chat~\cite{zhang2023next} prompts the LMM to output location embeddings, which are then decoded through a box decoder and a mask decoder to support a range of captioning and grounding tasks. Similarly, u-LLaVA~\cite{xu2024u} incorporates both a pixel-level decoder and a region-level decoder, along with corresponding projectors, to enable multi-level visual understanding. 

Subsequent works further advance toward handling more complex entities and employing a broader range of decoders. 
Lumen~\cite{jiao2024lumen} introduces a peak point selection mechanism that parses the heatmap into a set of points, each representing the center of an identified object or keypoint. These points are then processed through a box decoder and a promptable mask decoder, enabling task-specific decoding.
VisionLLM v2~\cite{wu2024visionllm} introduces a  super link mechanism that enables flexible information flow between the MLLM and various task-specific decoders. This design unifies visual perception, understanding, and generation within a single, cohesive framework.
Vitron~\cite{fei2024vitron} extends the multi-decoder architecture to both images and videos, enabling an LLM-to-decoder instruction-passing mechanism that operates over both discrete textual inputs and continuous signal embeddings.
REF-VLM~\cite{tai2025ref} boosts multi-task performance by integrating Mask-Guided Aggregation, a Latent Embeddings Router, and Parallel Group Hungarian Matching. It further introduces a Triplet-Based Referring Paradigm, which decouples key visual decoding dimensions using symbolic delimiters to enhance output structure and interpretability.

\subsection{Specialized Decoding Strategies}
\label{subsec:specialized-decoding}

This line of work focuses on task-specific decoding strategies designed for particular perception challenges. Here, decoding refers to the process in which the MLLM generates structured text outputs that are either directly mapped to task-specific results—such as segmentation masks, bounding boxes, or object relations (\S\ref{subsubsec:direct-structured-output}), or used as prompts to drive dedicated downstream vision modules (\S\ref{subsubsec:prompt-guided-downstream}). We summarize representative approaches in both directions as follows.

\subsubsection{Direct Structured Output for Task Completion}
\label{subsubsec:direct-structured-output}

Several methods leverage the MLLM’s ability to directly output structured results for vision tasks.
BuboGPT~\cite{zhao2023bubogpt} addresses grounding by instructing the LLM to output region captions in text, which are then matched to entities detected by RAM~\cite{zhang2024recognize} and Grounding DINO~\cite{liu2024grounding}; a subsequent entity-matching module aligns the textual outputs with semantic regions in the image.
LLaVA-Grounding~\cite{zhang2024llava} also focuses on grounding, directly projecting the language form response into a grounding model to produce bounding boxes.
ASMv2~\cite{wang2024all} further explores region-level and relation-aware predictions, where the MLLM is tasked with outputting object coordinates and predicting inter-object relations using special $\langle pre \rangle$ 
tokens.

For segmentation,
LLaFS~\cite{zhu2024llafs} enables end-to-end few-shot segmentation by letting the LLM produce segmentation polygons and constructing a region-attribute table to provide multi-modal guidance simulating human visual mechanisms.
LlamaSeg~\cite{deng2025llamaseg} reformulates image segmentation as a visual generation problem, representing masks as visual tokens and employing a LLaMA-style~\cite{touvron2023llama} Transformer to predict them directly from image
inputs.
VistaLLM~\cite{pramanick2024jack} introduces an instruction-guided image tokenizer, unifying multi-level vision-language tasks within a general-purpose framework. Specifically for segmentation, it employs a gradient-aware adaptive sampling technique to efficiently represent segmentation masks as sequences of points.
UFO~\cite{tang2025ufo} unifies diverse fine-grained perception tasks within the same open-ended language interface as vision-language tasks, without relying on task-specific decoders. Additionally, by reformulating segmentation as an embedding retrieval problem, it flexibly supports mask prediction within this unified framework.
RAS~\cite{cao2025refer} addresses omnimodal referring expression segmentation by first generating candidate mask groups with a segmentation model, and then tasking the mask-centric LMM to select the appropriate mask from this pool according to vision-language prompts.
HiMTok~\cite{wang2025himtok} introduces an efficient hierarchical mask tokenizer that represents mask images as hierarchical tokens, which adopts a three-stage training procedure and eliminates the need for any  external segmentation foundation model.

\subsubsection{Guiding Downstream Modules via Structured Prompts}
\label{subsubsec:prompt-guided-downstream}

Other works exploit the MLLM’s structured text outputs as prompts or intermediate guidance for external modules.
SeSaMe~\cite{wu2024see} adopts a cascading approach for the False Premise Correction task, splitting the process into “see”, “say”, and “segment” stages—where the first two stages generate language prompts that drive LISA~\cite{lai2024lisa} for final segmentation.
LLM-Seg~\cite{wang2024llm} reformulates segmentation as a selection task: the LLM, in tandem with mask proposals generated by SAM and DINO, produces $\langle seg \rangle$ tokens that interact with a fusion module and MLP for threshold-based mask selection.
HRSeg~\cite{lin2025hrseg} further extends LLM-Seg~\cite{wang2024llm} to high-resolution settings by introducing two additional modules.
SAM4MLLM~\cite{chen2024sam4mllm} tackles referring expression segmentation by letting the LLM predict prompt points for SAM, thus achieving segmentation without modifying the original MLLM architecture.
GroundedVideoLLM~\cite{wang2024grounded} extends this idea to the temporal domain, introducing special temporal tokens that share an embedding space with the LLM and employing a segment-wise encoding strategy for temporal awareness.
Text4Seg~\cite{lan2024text4seg} introduces text-as-mask paradigm. It encodes segmentation masks as semantic descriptors—a textual sequence—allowing the LLM to autoregressively predict the class for each image patch, with SAM as mask refiner.
READ~\cite{qian2025reasoning} introduces the Similarity as Points (SasP) module, in which $\langle seg \rangle$ and image tokens are used to compute similarity scores that identify salient locations in the image. These selected points are encoded as sparse embeddings, which are then transformed into continuous attention maps using Gaussian-weighted interpolation, guiding the model toward more precise segmentation and reasoning.

\section{Stage III: Dynamic Perception via Adaptive Processing}
\label{sec:stage3}
\begin{figure}[t]
    \centering
    \includegraphics[width=1\linewidth]{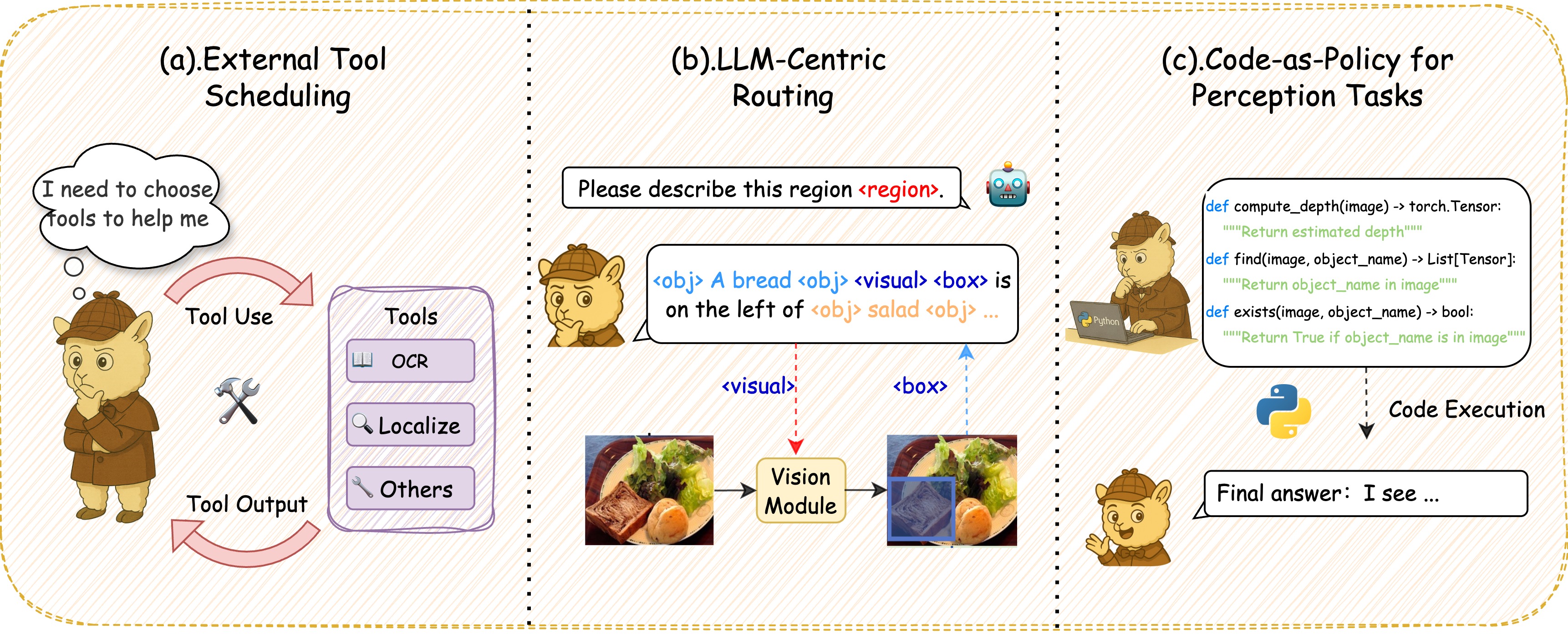}
    \caption{Dynamic Perception via Adaptive Processing.
(a) External Tool Scheduling: The LLM decides which vision tools to invoke (e.g., OCR, localization, etc.) and integrates their outputs back into its reasoning loop.
(b) LLM-Centric Routing: The LLM emits structured tags (e.g., visual token, box token) to direct specific image regions into corresponding vision modules, whose results are fed back for a final answer.
(c) Code-as-Policy for Perception Tasks: The LLM generates executable Python‐style functions (e.g., compute\_depth, find, exists), runs them on the image, and uses their outputs to produce the final response.}
    \label{fig:pic_dynamic_perception}
\end{figure}
Stage I and Stage II, whether through encoder-centric or decoder-centric modifications, have enabled MLLMs to achieve strong perceptual capabilities, progressing from region-level to pixel-level understanding. However, a core limitation of these approaches is their reliance on statically encoding image information in a single pass, restricting the amount and richness of visual input that the model can access.

Stage III marks a paradigm shift, introducing dynamic perception to MLLMs. Instead of static, fixed-pipeline processing, MLLMs now adopt adaptive, context-aware perception strategies. This new dynamic framework empowers MLLMs with greater flexibility and enhanced perceptual abilities across a diverse array of multimodal tasks.
As illustrated in Fig.~\ref{fig:pic_dynamic_perception}, we categorize dynamic perception methods into three major paradigms:
External Tool Scheduling (\S\ref{subsec:external-tool})
, LLM-Centric Routing (\S\ref{subsec:llm-centric-routing}), and Code-as-Policy (\S\ref{subsec:code-as-policy}). 

\subsection{External Tool Scheduling}
\label{subsec:external-tool}
This section introduces multi-stage tool scheduling, in which external tools are invoked and their outputs—returned in specific formats—are integrated to provide additional visual information, thereby enhancing the execution of vision-language tasks. We organize these works into two categories: Fixed Step Scheduling (\S\ref{subsubsec:Fixed Step Scheduling}) and Dynamic Expert Selection and Execution (\S\ref{subsubsec:Dynamic Expert Selection and Execution}).
\subsubsection{Fixed Step Scheduling}
\label{subsubsec:Fixed Step Scheduling}
Some early works decompose vision-language tasks into multiple fixed stages, leveraging chain-of-thought prompting~\cite{wei2022chain} or orchestrating multiple LLM agents to complete the task along a predefined workflow.
LLaVA-Plus~\cite{liu2024llava} extends LLaVA~\cite{liu2023visual} by integrating a large and diverse suite of external tools, which can be selectively composed and activated to solve real-world compositional tasks. The system supports API parameterization for invoking corresponding function arguments as needed. As a result, LLaVA-Plus demonstrates four major categories of individual skills—including understanding, external knowledge retrieval, visual prompting, and generation—as well as a variety of composed skills through flexible tool chaining and model orchestration. 

For task-specific methods, 
DetToolChain~\cite{wu2024dettoolchain} tackles zero-shot object detection with a detection prompting toolkit paradigm, where MLLMs are guided to perceive objects through region focusing, progressive prediction refinement, and contextual inference.
$P^2\text{G}$~\cite{chen2024plug} presents a plug-and-play framework for grounded reasoning in high-resolution, text-rich images. The agent adaptively decides whether to retrieve additional OCR or detection results as clues, and then fuses these clues into a multimodal prompt to support accurate, grounded reasoning with the MLLM.
ThinkFirst~\cite{kao2025think} introduces a training-free reasoning segmentation framework that uses CoT-guided prompts to analyze images, generating a detailed summary that, combined with the user query, directs the segmentation model.
VTPrompt~\cite{jiang2024joint} leverages GPT to extract key concepts from textual questions, which are then used to guide a detection model. The detection results are subsequently fed back to GPT for step-by-step reasoning, leading to improved zero-shot object-oriented perception performance.

\subsubsection{Dynamic Expert Selection and Execution}
\label{subsubsec:Dynamic Expert Selection and Execution}
Some methods implement dynamic perception and planning in reasoning by first selecting the appropriate expert or tool for each task, then followed by targeted execution. 
IoT~\cite{zhou2024image} uses Image-of-Thought prompting to let the MLLM autonomously plan and execute image processing steps, generating visual rationales that are combined with text to form comprehensive multimodal explanations. 
VipAct~\cite{zhang2024vipact} proposes an agent framework that enhances VLMs by integrating multi-agent collaboration and vision expert models. The framework features an orchestrator agent responsible for planning and coordination, a set of specialized agents for handling specific tasks, and vision expert models that deliver high-precision perceptual information. 
Similarly, TACO~\cite{ma2024taco} extends the CoT paradigm to chains-of-thought-and-action, leveraging a variety of pre-designed vision-centric and vision-language tools. During inference, the model executes intermediate steps by invoking these external tools, integrating both reasoning and action outputs to generate coherent responses.

\subsection{LLM-Centric Routing}
\label{subsec:llm-centric-routing}
The previous stage focused on scheduling, where the LLM served as a coordinator—whether following fixed-step pipelines or planning before selecting experts, the model did not actively engage in perception. In contrast, this stage emphasizes LLM-driven active routing of perception, where the LLM autonomously selects and routes perceptual processes based on context, enabling more proactive and dynamic control over multimodal information flow.
We elaborate in three parts: Step-by-step visual reasoning (\S\ref{subsubsec:Step-by-step visual reasoning}), Multi-round iterative visual search (\S\ref{subsubsec:Multi-round iterative visual search}) , and Detail enhancement via re-encoding (\S\ref{subsubsec:Detail enhancement via re-encoding}).

\subsubsection{Step-by-step visual reasoning}
\label{subsubsec:Step-by-step visual reasoning}
Some studies focus on step-by-step visual understanding by employing chain-of-thought-like reasoning.

Focus-and-zoom reasoning on a single image progressively narrows attention:
SEAL~\cite{wu2024v} integrates an MLLM with a localization module composed of an image backbone, a target localization decoder, and a search cue localization decoder. The MLLM generates additional $\langle loc \rangle$ tokens, which are fed to the respective decoders to produce target coordinates and search cue heatmaps. The system iteratively performs visual search until the desired criteria are met. 
DualFocus~\cite{cao2024dualfocus} decomposes the visual reasoning process into macro and micro perspectives. The model first examines the image from a macro perspective to answer the question and then identifies relevant sub-regions for focused micro-level analysis.
CoS~\cite{liu2024chain} introduces Chain-of-Spot that enhances feature extraction by interactively focusing on key regions of interest within the image according to the given instructions.
CoReS~\cite{bao2024cores} introduces the Chains of Reasoning and Segmenting through a dual-chain structure that generates multi-modal, chain-like outputs to aid the segmentation process.
TextCoT~\cite{luan2024textcot} decomposes text-rich image understanding into three stages: Image Overview, Coarse Localization, and Fine-grained Observation. The first two stages generate a global context description and identify an answer region, which is then cropped and sent back to the LLM for further analysis, ultimately enabling a more accurate final response.

Action-centric chains drive progress through explicit visual operations or object-level structure: 
CogCoM~\cite{qi2025cogcom} proposes Chain of Manipulations, that enables VLMs to solve problems step-by-step by actively manipulating visual inputs as evidence without relying on external tools. 
VolCano~\cite{li2024vocot} proposes a multi-step, visually- grounded, object-centric chain-of-thought reasoning framework. This approach constructs object-centric reasoning paths based on shared object-level information across modalities, while also providing visually grounded representations of object concepts through interleaved and aligned multimodal features.

Beyond single image, CMMCoT~\cite{zhang2025cmmcot} targets multi-image understanding and mimics human-like slow thinking by integrating coordinate-guided visual token extraction and a Retrieval-based Image Feature Reasoning Enhancement Module. It mitigates error accumulation and enhances cross-image visual concept tracking.

\subsubsection{Multi-round iterative visual search}
\label{subsubsec:Multi-round iterative visual search}
Other works emphasize decomposing perception into multiple rounds of visual search, progressively refining the analysis through iterative exploration. 
ZoomEye~\cite{shen2024zoomeye} introduces a tree search algorithm designed to navigate the hierarchical and visual nature of images to capture relevant information, enabling MLLMs to simulate human zooming actions by searching along the image tree.
Similarly, DyFo~\cite{li2025dyfo} introduces a training-free dynamic visual search approach based on Monte Carlo Tree Search, enabling the model to navigate visual spaces using focus nodes that amplify critical information and filter out irrelevant input. 
Zoom-Refine~\cite{yu2025zoom} targets high-resolution image understanding by combining MLLM-guided localized zoom—focusing on relevant details—with explicit self-refinement, where the model critically re-evaluates its initial assessment using the high-resolution crop as evidence against the broader context.

\subsubsection{Detail enhancement via re-encoding}
\label{subsubsec:Detail enhancement via re-encoding}
A third line of research aims to extract fine-grained details within a single dialogue turn by leveraging re-encoding or re-aligning  mechanisms for enhanced visual comprehension. 
For re-encoding strategies, 
VPT~\cite{yu2025introducing} enables MLLMs to autonomously control visual perception by introducing special $\langle ctr \rangle$ tokens, which guide the re-encoding of visual information for adaptive and task-driven processing.
Argus~\cite{man2025argus} incorporates a grounding-driven visual attention re-engagement mechanism that leverages object-centric grounding as visual chain-of-thought signals. By conditioning on multimodal input instructions, the model dynamically grounds the most relevant regions of interest, enabling more effective goal-directed visual attention during complex multimodal reasoning tasks.
COVLM~\cite{li2023covlm} introduces a set of communication tokens that enable the LLM to dynamically interact with both the vision encoder and detection network during decoding. By generating communication tokens alongside visual entities and relations, the LLM can request relevant regions from the detection network, which are then fed back for more context-aware language generation. 
LLaVA-Aurora~\cite{bigverdi2025perception} advances this paradigm by distilling fine-grained visual perception into MLLMs: the model learns to predict discrete Perception Token sequences—generated by a VQVAE—directly from images as intermediate reasoning steps, which are then incorporated into multi-task training.

As for re-aligning methods, ClawMachine~\cite{ma2024clawmachine} unifies diverse fine-grained referential tasks by directly annotating visual entities with corresponding token collectives, thus removing the need for additional syntactic structures and seamlessly integrating visual and language tokens within a unified auto-regressive sequence.
LIRA~\cite{li2025lira} equips a comprehension-based LMM with segmentation ability by employing a Semantic-Enhanced Feature Extractor, which performs dynamic cropping to capture fine-grained image features, and by aligning these representations with textual descriptions via an interleaved local visual coupling module.
VGR~\cite{wang2025vgr} introduces a selective feature replay module, which enables the MLLM to selectively attend to informative regions and retrieve image tokens from a memory pool, thereby enriching visual clues for reasoning.

\subsection{Code-as-Policy for Perception Tasks}
\label{subsec:code-as-policy}
A small number of works employ language model generated programs, executing code to accomplish perception tasks.

VisPorg~\cite{gupta2023visual} uses the in-context learning ability of large language models to generate Python-like modular programs. Each line of the generated program invokes one of several off-the-shelf computer vision models, image processing subroutines, or python functions to produce intermediate outputs that may be consumed by subsequent parts of the program.
Similarly, ViperGPT~\cite{suris2023vipergpt} introduces a simple framework for solving complex visual queries by integrating code-generation models into vision with an API and the Python interpreter.
VisualSketchpad~\cite{hu2024visual} introduces a framework that equips MLLMs with the ability to generate intermediate sketches via Python code, enabling collaboration with vision specialists for visual reasoning tasks.
Recently, PyVision~\cite{zhao2025pyvision} takes advantage of the advanced coding and multimodal understanding capabilities of modern MLLMs~\cite{claude4}, generating Python code to build and run complex tools, which enables more general and flexible reasoning.

For task-specific methods, 
LogiCode~\cite{zhang2024logicode} targets industrial logical anomaly detection by integrating LLMs for code generation and logical reasoning.
ReFocus~\cite{fu2025refocus} targets structured image understanding by enhancing MLLMs with visual reasoning as an intermediate step and providing an interface for generating visual artifacts using Python-based image editing tools.

\section{Stage IVA: Architecture-Free Strategies for Perception Enhancement}
\label{sec:stage4a}

\begin{figure}[t]
    \centering
    \includegraphics[width=1\linewidth]{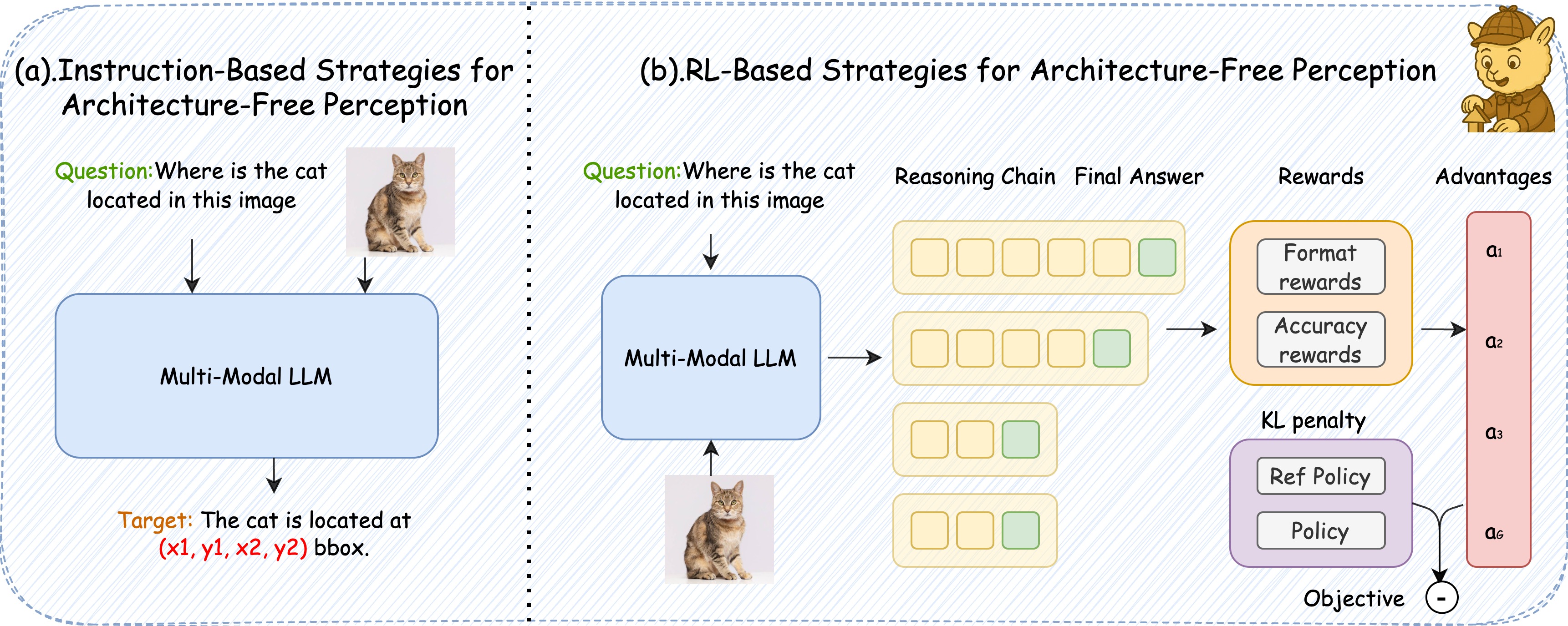}
    \caption{Architecture-Free Strategies for Perception Enhancement.  
(a) Instruction-Based Strategies: the LLM is guided by task-specific instructions to directly generate structured outputs (e.g., bounding box coordinates), without relying on explicit architectural modules.  
(b) RL-Based Strategies: the LLM produces multiple reasoning chains and candidate answers, which are optimized via reward signals that reflect task correctness and output format; a KL penalty against a reference policy helps maintain response quality during learning.}
    \label{fig:pic_instruction_rl}
\end{figure}
After Stage III, the perceptual capabilities of MLLMs have been greatly enhanced. Subsequent developments diverge into two main branches. One branch focuses on further improving performance for specific tasks without modifying the model architecture, primarily through advanced training strategies—most notably, the adoption of reinforcement learning methods (\S\ref{sec:stage4a}). The other branch moves beyond task-specific optimization, aiming instead to develop robust and versatile MLLMs with comprehensive perceptual abilities by integrating multiple approaches in a unified framework(\S\ref{sec:stage4b}).

In this stage, we focus on approaches that enhance perception capabilities without modifying the model architecture. As illustrated in Fig.~\ref{fig:pic_instruction_rl}, we categorize these works into Instruction-Based Strategies for Architecture-Free Perception (\S\ref{subsec:instruction-based}) and RL-Based Strategies for Architecture-Free Perception (\S\ref{subsec:rl-based}).

\subsection{Instruction-Based Strategies for Architecture-Free Perception}
\label{subsec:instruction-based}

In this subsection, we review early instruction-based methods that enhance perception capabilities through instruction tuning, enabling models to achieve finer-grained perceptual performance without modifying their underlying architecture.

A number of early representative works laid the foundation for instruction-based, architecture-free perception in MLLMs.
LLaVA~\cite{liu2023visual} serves as a pioneering work, utilizing language-only GPT-4 to generate multimodal instruction-following data. This enables the subsequent training of LLaVA, a multimodal model adept at following human intent for diverse visual tasks.
Kosmos-2~\cite{peng2023kosmos} further enhances multimodal grounding by introducing a novel training data format: it encodes bounding box coordinates as location tokens, which are appended to the corresponding text spans. This design acts as a hyperlink in the training data, seamlessly linking image regions to their textual descriptions and facilitating more effective vision-language alignment during model training.
Similarly, Shikra~\cite{chen2023shikra} and Pink~\cite{xuan2024pink} support spatial coordinate input and output by directly incorporating bounding box information into the question-answering process, enabling natural language interaction with spatially grounded content.
Apart from that, VPD~\cite{hu2024visualprogram} synthesizes training data for VLMs by generating programs that leverage external specialist models and tools. These LLM-generated programs are then used within an instruction tuning framework to distill cross-modal reasoning abilities and expert skills into multimodal models.

Several works further improve task-specific performance through tailored instruction tuning strategies.
VTimeLLM~\cite{huang2024vtimellm} targets fine-grained video moment understanding and reasoning by introducing a three-stage, temporally aware training framework.
GroundingGPT~\cite{li2024groundinggpt} addresses multi-modal grounding with a three-stage, coarse-to-fine training strategy, supported by stage-specific datasets to effectively enhance grounding capabilities.
LocVLM~\cite{ranasinghe2024learning} focuses on spatial reasoning by encoding image coordinates into language and proposing three instruction fine-tuning  objectives, thereby equipping MLLMs with the ability to reason about spatial composition through embedded image coordinates in text.
Migician~\cite{li2025migician} focuses on multi-image grounding by constructing a large-scale dataset for training across six representative tasks, and then fine-tuning the model on free-form instruction data.

\subsection{RL-Based Strategies for Architecture-Free Perception}
\label{subsec:rl-based}

Recently, following the introduction of OpenAI-O1~\cite{jaech2024openai} and DeepSeek-R1~\cite{guo2025deepseek}, a series of studies have explored the application of reinforcement learning (RL) in the vision-language domain, leading to notable progress in architecture-free perception enhancement. We elaborate in R1-Style Reinforcement Learning (\S\ref{subsubsec:R1-Style Reinforcement Learning}) and Other Reinforcement Learning Methods (\S\ref{subsubsec:Other Reinforcement learning methods}).

\subsubsection{R1-Style Reinforcement Learning}
\label{subsubsec:R1-Style Reinforcement Learning}

Following the success of DeepSeek’s work~\cite{shao2024deepseekmath,guo2025deepseek}, several studies have introduced Group Relative Policy Optimization (GRPO)~\cite{shao2024deepseekmath} into visual tasks, employing task-specific, rule-based reward design to provide stable and interpretable reward signals. This approach has notably improved the performance of MLLMs on specific visual tasks. 

Among these, VLM-R1~\cite{shen2025vlm} and Seg-Zero~\cite{liu2025seg} stand out as representative works, focusing respectively on detection and segmentation tasks in computer vision.
VLM-R1~\cite{shen2025vlm} introduces R1-style reinforcement learning into the domains of referring expression comprehension and open-vocabulary object detection, building on open-r1~\cite{faceopen}. By incorporating dedicated accuracy and format rewards, VLM-R1 demonstrates substantial gains in both task performance and out-of-domain generalization.
Seg-Zero~\cite{liu2025seg} targets reasoning segmentation and adopts a decoupled architecture consisting of a reasoning model and a segmentation model. By employing a purely reinforcement learning algorithm, Seg-Zero exhibits emergent reasoning abilities.

Building on these seeds, a line of work extends verifiable-reward RL to a wider set of static  vision-language tasks. 
R1-V~\cite{chen2025r1v} addresses object-counting using GRPO, while Visual-RFT~\cite{liu2025visual} and DIP-R1~\cite{park2025dip} apply GRPO-based reinforcement learning to enhance visual perception and grounding capabilities. 
UniVG-R1~\cite{bai2025univg} generalizes this approach to universal visual grounding while Perception-R1~\cite{yu2025perception} extends it to several representative downstream perception tasks. 
Vision-R1~\cite{zhan2025vision} proposes a vision criterion-driven reward and progressive rule refinement for improved object localization. 
Ground-R1~\cite{cao2025ground} targets grounded visual reasoning and decouples evidence region generation from answer synthesis, enabling interpretable reasoning via format-constrained grounding and reward-driven response generation.
Rex-Thinker~\cite{jiang2025rex} reformulates grounded object referring as a planning–action–summarization problem to achieve grounded and interpretable predictions. The model first detects candidate objects and then performs step-by-step verification against the referring expression through a structured planning, action, and summarization reasoning process. 
Zhang et al.\cite{zhang2025improving} further extends this reasoning ability to multi-image grounding task 
while Curr-ReFT~\cite{deng2025boosting} extends both reasoning and OOD generalization capabilities to small-scale MLLMs.
GRIT~\cite{fan2025grit} enables models to generate visually grounded reasoning chains by interleaving natural language with explicit bounding box coordinates that reference relevant image regions, utilizing a specially designed GRPO algorithm.
Similarly, SATORI~\cite{shen2025satori} decomposes VQA into three verifiable stages—global image captioning, region localization, and answer prediction—to provide step-by-step explicit reward signals.
Griffon-R~\cite{zhan2025understand} also adopts a staged approach, decomposing the reasoning process into understanding, thinking, and answering.
Likewise, TreeVGR~\cite{wang2025traceable} introduces explicit bounding boxes into the reasoning chain and assigns a Traceable Evidence Reward based on box IoU throughout the process. DeepPerception~\cite{ma2025deepperception} demonstrates that integrating cognitive visual perception—enhanced by external knowledge and reasoning—significantly boosts fine-grained perception in MLLMs. 
Relation-R1~\cite{li2025relation} extends these verifiable-reward GRPO methods to N-ary relation detection within a unified relation comprehension framework.

Another line of works focus on segmentation-centric advances and evidence-traceable rewards. On the segmentation front, PixelThink~\cite{wang2025pixelthink} introduces an efficiency-aware reasoning scheme built on Seg-Zero~\cite{liu2025seg} that dynamically regulates reasoning length based on task difficulty and model uncertainty.
Seg-R1~\cite{you2025seg} and SAM-R1~\cite{huang2025sam} incorporate GRPO into generating point and bounding box prompts in
the next-token fashion, which are then used to guide SAM2~\cite{ravi2024sam} in producing segmentation masks.
VisionReasoner~\cite{liu2025visionreasoner} introduces a unified framework for reasoning across diverse visual perception tasks, leveraging tailored reward functions and training strategies to enable comprehensive multi-task visual reasoning within a single model. ALToLLM~\cite{wang2025alto} incorporates an adaptive-length mask tokenizer into the MLLM, allowing adaptive mask token generation for object segmentation tasks, with GRPO applied during training. 

For anomaly detection, Anomaly-R1~\cite{chao2025anomalyr1} and OmniAD~\cite{zhao2025omniad} leverage GRPO to enhance industrial anomaly detection performance, while LAD-Reasoner~\cite{li2025lad} extends conventional anomaly detection by incorporating logical reasoning with a human-interpretable reasoning process.

In the video domain and multi-image  reasoning, DeepVideo-R1~\cite{park2025deepvideo} employs regressive GRPO and difficulty-aware data augmentation to address complex video reasoning challenges. STAR-R1~\cite{li2025star} adopts a single-stage pure RL paradigm for transformation-driven reasoning, MiCo~\cite{chen2025mico} leverages inherent image constraints to incentivize multi-image reasoning in MLLMs, and VideoCap-R1~\cite{meng2025videocap} utilizes structured thinking and dual reward mechanisms for nuanced video captioning. VersaVid-R1~\cite{chen2025versavid} further extends the capabilities of MLLMs to multiple-choice and open-ended question answering tasks within the Reason-Then-Respond paradigm.
AvatarShield~\cite{xu2025avatarshield} targets Human-Centric Video Forgery Detection and integrates GRPO into the training.
VAU-R1~\cite{zhu2025vau} further demonstrates the effectiveness of R1-style reinforcement learning.

\subsubsection{Other Reinforcement Learning Methods}
\label{subsubsec:Other Reinforcement learning methods}
Beyond R1-style frameworks, other works explore alternative reinforcement learning approaches to optimize model performance on specific vision-language tasks.
POPEN~\cite{zhu2025popen} targets reasoning segmentation by introducing a novel preference-based optimization approach and leveraging ensemble methods with task-specific designs.
PerPO~\cite{zhu2025perpo} introduces Perceptual Preference Optimization to align with the human perception process for discriminative tasks.
VideoChat-TPO~\cite{yan2025task} proposes Task Preference Optimization and enhances fine-grained spatio-temporal perception in videos by introducing task-specific heads. 
VisReP~\cite{khan2024self} treats the LLM as a policy and applies reinforced self-training, utilizing feedback from visual program execution to progressively improve the model’s visual program synthesis abilities.
Insight-V~\cite{dong2025insight} builds a multi-agent system that decomposes visual reasoning
tasks into reasoning and summarization with a two-stage training pipeline.
VisRL~\cite{chen2025visrl} applies reinforcement learning to intention-driven visual perception tasks by leveraging self-generated data and self-assigned rewards. The model is iteratively updated using step-level Direct Preference Optimization (DPO)~\cite{rafailov2023direct}, establishing a learning process that is much closer to human-like visual understanding.
SegAgent~\cite{zhu2025segagent} reformulates segmentation as a multi-step decision process, imitating human annotation trajectories with agent-based models and fine-tuning MLLMs on these paths.

\section{Stage IVB: Towards Unified Perception — A Perspective on the Convergence of Instruction, Adaptivity, and RL}
\label{sec:stage4b}
\begin{figure}[t]
    \centering
    \includegraphics[width=1\linewidth]{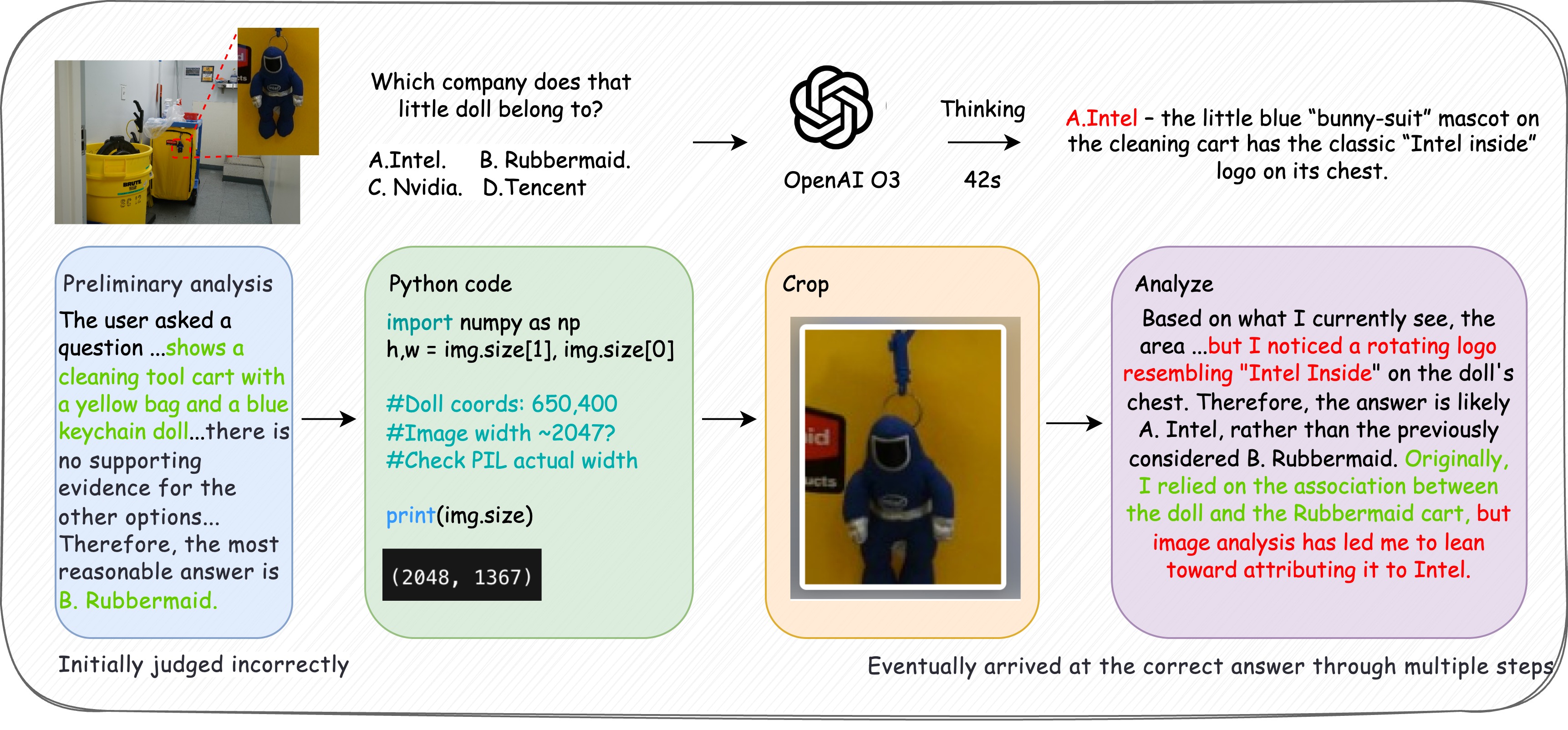}
    \caption{Case study of OpenAI o3’s thinking with images, reaching the correct answer after 42 seconds of reasoning. The question is from V* Bench.}
    \label{fig:pic_case_study}
\end{figure}
Stage IVB builds on all previous stages and takes a further step by integrating more complex capabilities and control mechanisms. Rather than adhering to the external tool scheduling view of Stage III, it instead aims for a more unified yet flexible perception–reasoning paradigm, in which diverse perceptual operations and tool calls can be composed and reused across tasks.

We organize this chapter as follows: we first provide a brief review and analysis of the evolution of overall structural paradigms, from architectural modifications to collaborative paradigm shifts, while also clarifying the limitations inherent in each development stage (see \S\ref{subsec:why-unification-matters}). Next, we examine recent research directions that build upon these foundations by integrating instruction tuning, adaptivity, and reinforcement learning, which have driven notable progress in the field (see \S\ref{subsec:signs-of-convergence}). Finally, we look ahead to the future, discussing both the outstanding challenges and the promising opportunities for the next generation of multimodal perception agents (see \S\ref{subsec:new-gen-perception-agents}).

\subsection{Why Unification Matters: The Need for a Converged Paradigm}
\label{subsec:why-unification-matters}
In this subsection, we begin with a brief recap of the key advances and limitations at each stage (\S\ref{subsubsec:A Brief Retrospect}), and then empirically propose three essential criteria that are critical for achieving true unification in vision-language perception (\S\ref{subsubsec:The Imperative for Unification}).

\subsubsection{A Brief Retrospect}
\label{subsubsec:A Brief Retrospect}
A retrospective analysis of Stage I (encoder-centric) and Stage II (decoder-centric) reveals a clear progression in perceptual granularity, advancing from image-level to region-level and ultimately to pixel-level understanding. However, these advances are still fundamentally limited by their reliance on one-shot, static encoding of visual information. Stage III introduces a shift toward dynamic and adaptive perception, yet the underlying scheduling strategies are largely heuristic or reliant on external tools, lacking a true closed loop with the model’s internal representations and optimization processes. Stage IVA further pushes the boundaries by leveraging reinforcement learning and verifiable rewards to improve performance without altering model architecture. Nevertheless, most of these approaches are still confined to isolated or fixed task types—such as detection, segmentation, or counting—and fall short of achieving a unified perception–reasoning cycle in open-ended scenarios.

\subsubsection{The Imperative for Unification}
\label{subsubsec:The Imperative for Unification}
Given these limitations, we argue that the next phase must move beyond structural modifications and collaborative designs toward a genuinely unified paradigm for perception. Such a paradigm should at least satisfy the following criteria:
\begin{itemize}
    \item \textbf{Unified intermediate representation}: Capable of naturally bridging text, regions, and pixels;
    \item \textbf{Unified task interface}: Replacing task-specific heads with structured outputs, programmatic actions, or verifiable rewards that reinterpret each task as an instance of a shared perception–reasoning cycle;
    \item \textbf{Unified resource-awareness}: Incorporating cost-aware perception–reasoning scheduling.
\end{itemize}

As a natural next step, we turn to the emergent signs of convergence already present in the literature, where instruction tuning, adaptive perception, and reinforcement learning are being integrated into closed-loop frameworks. These early explorations point the way toward truly unified paradigms for  vision-language perception.

\subsection{Signs of Convergence: Early Explorations in Unification}
\label{subsec:signs-of-convergence}
Recently, a handful of pioneering studies have emerged, with the OpenAI o3~\cite{openai2025thinkingwithimages} model representing one of the most notable advances. Although the technical details and training pipeline of o3 have not been released, both technical blog analyses and our empirical observations suggest that o3 tackles complex perception-driven tasks by combining planning, dynamic tool selection and usage (e.g., zoom-in, crop, contrast adjustment), Python code execution, and internet search capabilities. As shown in Fig.~\ref{fig:pic_case_study}, o3 provides a compelling case of unified vision-language perception integrating reasoning, external tool orchestration, and adaptive processing. 

\subsubsection{Zoom-In and Attention-Focused Strategies}
\label{subsubsec:zoom-in-strategies}
Several studies have attempted to replicate aspects of o3, with a subset specifically centering on zoom-in and adaptive attention mechanisms. Notably, Active-o3~\cite{zhu2025active} adopts a GRPO-based reinforcement learning framework that empowers MLLMs with active perception. Through a dual-policy design for sensing and action, and by optimizing with task-aware and exploratory rewards, Active-o3 enables MLLMs to reason about where to attend and how to interact with visual input more effectively,  closely mirroring o3’s zoom-in search strategy. 
Similarly, Kumar et al.~\cite{kumar2025reinforcing} extend this approach to resource-constrained settings, focusing on small-scale MLLMs with limited parameter counts. 
DeepEyes~\cite{zheng2025deepeyes} empowers MLLMs with native “thinking with images” capabilities through end-to-end reinforcement learning, seamlessly blending visual inputs with textual reasoning and forming iMCoT with a built-in zoom-in tool—all without requiring cold-start supervised fine-tuning or relying on separate specialized models as external tools. 
CoF~\cite{zhang2025chain} utilizes a Chain-of-Focus method, enabling MLLMs to adaptively focus and zoom in on key image regions based on visual cues and given questions, through a two-stage training pipeline.
PixelReasoner~\cite{su2025pixel} employs a two-stage approach—using instruction tuning to learn visual operations, then reinforcement learning with a curiosity-driven reward to balance pixel-space and textual reasoning.
In contrast to the above methods that explicitly inject visual cues, Look-Back~\cite{yang2025look} enables models to autonomously decide when and how to refocus on visual inputs by emitting special ⟨back⟩ tokens, allowing them to re-evaluate and correct their reasoning against the image.

\subsubsection{Adaptive Tool-Use Frameworks}
\label{subsubsec:tool-use-frameworks}
In addition to zoom-in strategies inspired by o3, another group of works extends the o3 paradigm by focusing on adaptive tool selection and orchestration. 
OpenThinkIMG~\cite{su2025openthinkimg} introduces an end-to-end tool-augmented MLLM framework, anchored by V-ToolRL—a reinforcement learning approach that enables adaptive and efficient tool selection via direct interaction and reward feedback, surpassing static trajectory imitation.
VisTA~\cite{huang2025visualtoolagent} introduces a reinforcement learning framework that enables visual agents to autonomously select effective external tools, learning adaptive tool-selection strategies without explicit supervision.
VRAG-RL~\cite{wang2025vrag} introduces a framework for training MLLMs to reason about, retrieve, and understand visually rich information via a visual perception action space—including selection, cropping, and scaling of regions of interest—and a comprehensive reward mechanism.

\subsubsection{Advanced Reasoning and Refinement Techniques}
\label{subsubsec:advanced-refinements}
Beyond zoom-in and tool-use frameworks, several studies introduce additional mechanisms and innovations. 
VisionThink~\cite{yang2025visionthink} leverages the LLM-as-Judge strategy and a tailored reward function for General VQA that enhances efficiency and performance, by initially processing a downsampled image and using reinforcement learning to selectively upscale to higher resolution when needed. 
GThinker~\cite{zhan2025gthinker} proposes a cue-guided rethinking mechanism, enabling the model to move beyond rigid templates. This design supports flexible, question-driven reasoning and allows for robust handling of imperfect visual cues through reflective and knowledge-grounded thinking.
Open-Vision-Reasoner~\cite{wei2025open} extends prior linguistic cognitive behaviors into the visual domain, introducing visual reflection, divide-and-conquer, visual verification, and goal-driven visual tracing. The framework adopts a lightweight Proximal Policy Optimization (PPO)~\cite{schulman2017proximal} algorithm with Generalized Advantage Estimation (GAE)~\cite{schulman2015high} to enable effective and efficient visual reasoning.
\subsection{Towards a New Generation of Perception-Centric Agents: Opportunities and Challenges}
\label{subsec:new-gen-perception-agents}

% {\color{blue}
Despite significant advancements in the multimodal perception capabilities of MLLMs, the field is simultaneously confronting unprecedented challenges. In this section, we first address three critical scientific issues that have emerged.

\textbf{The Limitations of Conventional Evaluation}.
Traditional datasets are increasingly struggling to meet current research demands. On one hand, proxy tasks such as static detection and segmentation are often \emph{limited by the precision of human labeling}, making it difficult to accurately measure the nuanced performance of MLLMs~\cite{cao2024introducing}. On the other hand, these datasets frequently \emph{remain decoupled from practical downstream applications} ~\cite{qiang2025ver, jiang2025mac}. Furthermore, the performance of large models on general-purpose benchmarks has begun to \emph{reach a point of saturation}~\cite{li2025information}. This naturally leads to our first inquiry: \emph{How can we precisely define and evaluate the performance of MLLMs? Are traditional proxy tasks still valid, and to what extent do they truly represent real-world utility?}

\textbf{The "Generalization vs. Fine-tuning" Dilemma.} 
With the rapid iteration of foundational multimodal models—where major versions are updated approximately every year—the benefits of fine-tuning on medium-scale computer vision datasets (such as COCO~\cite{lin2014microsoft}) have become remarkably limited. In many cases, fine-tuning can \emph{severely compromise the generalization capabilities inherited from pre-training}~\cite{wu2025mitigating, hwang2026model,zhai2023investigating}. Frequently, researchers find that substantial \emph{performance gains can be achieved simply by waiting for the next iteration of a foundation model rather than through task-specific optimization}. As foundational capabilities improve, many legacy problems are resolved "automatically," posing a significant challenge for researchers in identifying meaningful long-term research directions.

\textbf{Redefining Multimodality and Representation.} 
From a broader perspective, the very definition of "Multimodal" remains a point of contention. Current research predominantly categorizes data by modality (vision, language, audio, etc.), which fosters a fragmented perspective. As emphasized in the preceding chapters of this survey, capabilities acquired solely from language-modality pre-training do not necessarily transfer seamlessly to other modalities~\cite{yu2026modality}. While current models achieve impressive results by scaling data—primarily through language pre-training supplemented by large-scale multimodal alignment—this scaling paradigm will inevitably encounter a bottleneck. Whether a truly unified representation space exists across all modalities remains a subject requiring deeper exploration~\cite{wang2024emu3, huh2024platonic}.

Furthermore, in the short term, MLLMs continue to face specific technical bottlenecks, which we will discuss in the following subsections.
%}
\subsubsection{Dependence on High-Quality Data Curation}

Most current efforts have focused on narrow, task-specific settings, and models remain far from achieving truly “general” perception. In practice, endowing models with more complex, general-purpose perceptual abilities requires integrating heterogeneous data sources~\cite{zheng2025deepeyes,zhan2025gthinker,wei2025open}. Moreover, when models incorporate additional operations—such as active zooming or multi-step interaction—these data demands become even more stringent~\cite{zhu2025active,zhang2025chain,su2025pixel}, often rendering the process prohibitively expensive and resource-intensive. Future work may  therefore explore more cost-effective strategies for scaling datasets~\cite{dong2025scalable} or investigate self-play–style paradigms~\cite{silver2017mastering,xiong2025llava} to reduce reliance on large-scale manual annotation.

\subsubsection{Lack of General and Fine-Grained Rewards}

As reinforcement‐learning–based approaches—particularly GRPO—have become more prevalent, model performance increasingly depends on carefully crafted reward functions. To date, most studies employ task‐specific reward schemes that, while well‐suited to their individual objectives, hinder MLLMs from offering a unified solution across diverse visual tasks. Moreover, GRPO itself suffers from intrinsic drawbacks, including the vanishing advantage problem in long‐horizon reasoning and stability issues~\cite{yu2025dapo,zheng2025group}. 
Besides, in contrast to the binary correctness metrics commonly used in language reasoning, perception tasks demand dense, fine‐grained supervision. Looking ahead, we advocate for the development of reward frameworks that are (1) task‐agnostic~\cite{wang2025unified}, enabling broad applicability; and (2) hierarchically fine‐grained\cite{wang2025towards}, providing supervision at multiple levels of abstraction. Such designs will be critical for advancing robust, unified vision–language perception in open‐world settings.

\subsubsection{High Computation Costs}
\label{subsubsec:efficient-mllms}
Recent zoom-in and tool-scheduling approaches have delivered impressive accuracy gains, but at the cost of considerable compute. For example, DeepEyes~\cite{zheng2025deepeyes} and GThinker~\cite{zhan2025gthinker} both train 7B-parameter models on \textit{4 nodes × 8 GPUs}, while still conceding limitations: the authors of DeepEyes note that they “only utilized Qwen2.5-VL-7B, whose fundamental capability is constrained by its small size,” and that their current pipeline “supports only the \textit{crop} operation, whereas real scenarios demand a richer toolset such as web search or drawing auxiliary lines.” Similar resource hurdles—and the inability to integrate a broader repertoire of visual tools—are echoed across the literature. 
Closing this efficiency gap will require the development of more cost-effective training methods and more efficient model architectures~\cite{zheng2025group,chen2025minimax} .

\section{Conclusion}
\label{sec:conclusion}

In this survey, we present the first systematic review of unified vision-language perception in multimodal large language models. By defining perception as an intrinsically integrated vision-language capability, we systematically trace the evolution of perception paradigms in MLLMs across five key stages, from encoder-centric and decoder-centric optimizations to dynamic, adaptive processing and architecture-free strategies. We further discuss the convergence of instruction tuning, adaptivity, and reinforcement learning, highlighting their emerging role in shaping the next generation of perception-centric agents.
Despite rapid progress, significant challenges remain, including (i) heavy reliance on curated multi-task data, (ii) absence of general, fine-grained rewards, and (iii) prohibitive computational costs without effective cost-aware scheduling. We hope that our taxonomy and analysis will provide the community with a clearer roadmap, inspire new directions, and accelerate the development of MLLMs toward truly general, unified multimodal intelligence.

\section*{CRediT authorship contribution statement}
\textbf{Haoxiang Sun}: Conceptualization, Methodology, Formal analysis, Investigation, Writing - Original Draft, Visualization;
\textbf{Tao Wang}: Conceptualization, Methodology, Validation, Formal analysis, Writing - Review \& Editing, Funding acquisition, Project administration, Supervision;
\textbf{Li Yuan}: Writing - Review \& Editing, Supervision;
\textbf{Jian Zhao}: Writing - Review \& Editing, Supervision;
\textbf{Jiancheng Lv}: Resources, Writing - Review \& Editing, Funding acquisition.

\section*{Statement on AI Writing Assistance}
During the preparation of this work, the authors used ChatGPT to improve the clarity and correct grammatical errors in the manuscript. After using this tool, the authors carefully reviewed and edited the content as needed and take full responsibility for the content of the publication. Additionally, ChatGPT-4o was employed to generate visualizations for demonstration purposes.

\section*{Declaration of competing interest}
The authors declare that there are no known competing financial interests or personal relationships that could have appeared to influence the work reported in this paper.

\section*{Data availability}
This article did not generate new datasets or code. All datasets and tools discussed are publicly available; additional information can be obtained from the corresponding author upon reasonable request.

\section*{Acknowledgment}
This work is supported by the National Science Foundation of China under Grant 62506249, the National Major Scientific Instruments and Equipments Development Project of National Natural Science Foundation of China under Grant 62427820, the Natural Science Foundation of Sichuan under grant 2024NSFSC1462, and the Fundamental Research Funds for the Central Universities under grant YJ202342. 

\bibliographystyle{elsarticle-num-names} 
\bibliography{m_revised}  % 对应 refs.bib
% \begin{thebibliography}{00}

% %% For authoryear reference style
% %% \bibitem[Author(year)]{label}
% %% Text of bibliographic item

% \bibitem[Lamport(1994)]{lamport94}
%   Leslie Lamport,
%   \textit{\LaTeX: a document preparation system},
%   Addison Wesley, Massachusetts,
%   2nd edition,
%   1994.

% \end{thebibliography}
\end{document}